\documentclass{article} 
\usepackage{iclr2026_conference,times}

\usepackage{hyperref}       
\hypersetup{
    colorlinks=true,
    linkcolor=red,
    filecolor=magenta,
    urlcolor=cyan,
    citecolor=cyan,
}

\usepackage{url}            
\usepackage{cite}
\usepackage{booktabs}       
\usepackage{amsfonts}       
\usepackage{amsmath}
\usepackage{nicefrac}       
\usepackage{microtype}      
\usepackage{subcaption}
\usepackage{multirow}
\usepackage{enumitem}
\usepackage{colortbl}
\usepackage{graphicx}
\usepackage[most]{tcolorbox}

\title{Perception-R1: Advancing Multimodal Reasoning Capabilities of MLLMs via Visual Perception Reward}

\author{%
    \textbf{Tong Xiao}$^{1}$, \textbf{Xin Xu}$^{4}$, \textbf{Zhenya Huang}$^{1,2}$, \textbf{Hongyu Gao}$^{3}$,\\ \textbf{Quan Liu}$^{2,3}$, \textbf{Qi Liu}$^{1,2}$, \textbf{Enhong Chen}$^{1,2}$\thanks{Corresponding author.} \vspace{1mm} \\
    $^{1}$University of Science and Technology of China \\
    $^{2}$State Key Laboratory of Cognitive Intelligence \\
    $^{3}$iFLYTEK AI Research \\
    $^{4}$Hong Kong University of Science and Technology \\
}

\newcommand{\methodname}{{Perception-R1}}

\iclrfinalcopy 
\begin{document}

\maketitle

\begin{abstract}
Enhancing the multimodal reasoning capabilities of Multimodal Large Language Models (MLLMs) is a challenging task that has attracted increasing attention in the community. Recently, several studies have applied Reinforcement Learning with Verifiable Rewards (RLVR) to the multimodal domain in order to enhance the reasoning abilities of MLLMs. However, these works largely overlook the enhancement of multimodal perception capabilities in MLLMs, which serve as a core prerequisite and foundational component of complex multimodal reasoning. Through McNemar's test, we find that existing RLVR method fails to effectively enhance the multimodal perception capabilities of MLLMs, thereby limiting their further improvement in multimodal reasoning. To address this limitation, we propose {\methodname}, which introduces a novel visual perception reward that explicitly encourages MLLMs to perceive the visual content accurately, thereby can effectively incentivizing both their multimodal perception and reasoning capabilities. Specifically, we first collect textual visual annotations from the CoT trajectories of multimodal problems, which will serve as visual references for reward assignment. During RLVR training, we employ a judging LLM to assess the consistency between the visual annotations and the responses generated by MLLM, and assign the visual perception reward based on these consistency judgments. Extensive experiments on several multimodal math and general benchmarks demonstrate the effectiveness and robustness of our {\methodname}, which achieves superior performance on all benchmarks using only 1,442 training data.
Our code and dataset will be available at \url{https://github.com/tongxiao2002/Perception-R1}.
\end{abstract}

\section{Introduction}

Multimodal reasoning is a fundamental capability for AI systems to solve complex real-world tasks and represents a critical step toward artificial general intelligence (AGI).
Since the emergence of Multimodal Large Language Models (MLLMs), extensive studies have been proposed to enhance their multimodal reasoning capabilities.
Early efforts focused on prompt engineering techniques~\citep{zhang2023multimodal,zheng2023ddcot,zhang2024cocot,mitra2024compositional,xue2024decompose}, while subsequent works leveraged supervised fine-tuning (SFT) with high-quality reasoning datasets~\citep{gao2023g,shi2024math,peng2024multimath,zhang2024mavis,liu2024socraticlm}. More recently, the success of reasoning models such as OpenAI-o1~\citep{jaech2024openai}, DeepSeek-R1~\citep{guo2025deepseek}, and Kimi-k1.5~\citep{team2025kimi} has demonstrated the efficacy of large-scale reinforcement learning (RL), particularly RL with Verifiable Rewards (RLVR), in enhancing advanced reasoning capabilities of LLMs and MLLMs, thereby drawing significant attention from the research community.

Motivated by these pioneering works, a growing number of studies~\citep{meng2025mm,peng2025lmm,yang2025r1,huang2025vision,deng2025openvlthinker} have emerged to further advance the multimodal reasoning capabilities of MLLMs leveraging RLVR, which train MLLMs solely with rewards derived from the correctness of the final answer. For instance, MM-Eureka~\citep{meng2025mm} and R1-VL~\citep{zhang2025r1} successfully stabilized the training of RLVR in the multimodal domain by incorporating additional training tricks and mechanisms. R1-OneVision~\citep{yang2025r1} and Vision-R1~\citep{huang2025vision} first constructed a large-scale high-quality multimodal dataset by addressing the modality gap issue, then applied a training pipeline consisting of cold-start followed by RLVR to train MLLMs on the dataset. All these works effectively enhanced the multimodal reasoning capabilities of MLLMs, leading to substantial performance improvements.

Multimodal reasoning can be naturally decomposed into multimodal perception and logical reasoning~\citep{amizadeh2020neuro,zhou2024proreason}, where multimodal perception is responsible for accurately understanding the multimodal input and supplying essential information for subsequent reasoning, thereby serving as the foundation for effective multimodal reasoning.
Although RLVR-trained MLLMs demonstrate improved reasoning capabilities, our detailed analysis reveal that existing RLVR fails to effectively improve the multimodal perception capabilities of MLLMs, making it a major bottleneck that restricts their further advancement in multimodal reasoning.

\begin{figure}[t]
    \centering
    \includegraphics[width=\linewidth]{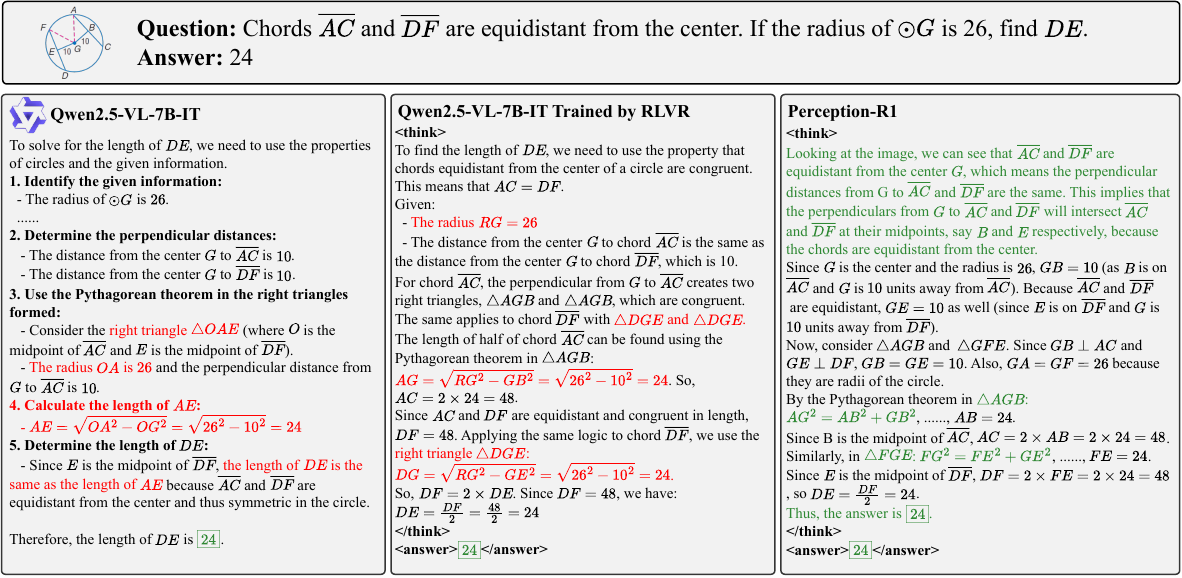}
    \caption{A comparison of three MLLMs on a geometry problem. Both Qwen2.5-VL-7B-IT and its RLVR-trained variant make severe perception errors but manage to guess the answer, whereas our {\methodname} first accurately describes the image and then solves the problem correctly.}
    \label{fig.introduction}
\end{figure}

For example, as illustrated in Figure~\ref{fig.introduction}, the original MLLM (left in Figure~\ref{fig.introduction}) makes severe multimodal perception errors (e.g., referring to ``right triangle $\triangle OAE$'' that does not exist in the image), indicating its limited multimodal perception capabilities. Nevertheless, it still manage to guess the correct answer. This makes existing RLVR method, which optimizes MLLMs solely based on answer accuracy, struggles to correct perception errors and may even reinforces this flawed reasoning path. Consequently, the resulting MLLM (middle in Figure~\ref{fig.introduction}) still exhibits weak multimodal perception capabilities similar to its original counterpart (e.g., referring to ``$RG$'' that does not exist), hindering the development of genuine multimodal reasoning capabilities.

We attribute this challenge to the rewards sparsity for multimodal perception when training MLLMs with existing RLVR, making it difficult to effectively enhance the multimodal perception capabilities of MLLMs. To address this challenge, we propose {\methodname}, which incorporates a novel and effective visual perception reward into the multimodal RLVR training process. The visual perception reward provides an additional reward signal beyond answer accuracy, explicitly encouraging MLLMs to perceive visual content accurately, thereby alleviating reward sparsity in RLVR training and facilitating more effective multimodal reasoning by strengthening the MLLMs' perceptual foundation.

Specifically, we introduce visual annotations into RLVR as auxiliary references, encouraging MLLMs to generate perception-accurate responses that closely align with them during training.
To obtain such visual annotations, we first collect CoT trajectories with correct final answers from a state-of-the-art multimodal reasoning model and then employ an LLM to extract natural language visual annotations from them. Our manual examination indicates that these visual annotations reach an accuracy of 96\% (see Appendix~\ref{app.sec.visual_examination}).
During RLVR training, visual perception reward is assigned based on the consistency between the visual annotations and the responses of MLLM, as evaluated by a judging LLM via prompting.
By incorporating the visual perception reward into RLVR, our {\methodname} achieves the best performance compared to several strong baselines across most multimodal benchmarks using only 1,442 training samples, surpassing Vision-R1~\citep{huang2025vision}, which requires 200K data samples for training in total.

In summary, our contributions are threefold:
\newline
(1). We investigate the behaviors of RLVR-trained MLLMs and their original counterparts, and find that their multimodal perception capabilities are not statistically significantly different, remaining a major bottleneck that limits further advancement in multimodal reasoning.
\newline
(2). We propose {\methodname}, which introduces a novel visual perception reward into RLVR. By providing an additional perception reward signal, {\methodname} alleviates the reward sparsity in multimodal perception and effectively enhances the multimodal reasoning capabilities of MLLMs.
(3). Extensive experiments on several multimodal math and general benchmarks demonstrate the superiority of our {\methodname}, which exhibits significantly improved multimodal perception capabilities and achieves superior performance on all benchmarks using only 1,442 training samples.

\section{Related Work}

\subsection{Multimodal Large Language Models}
Multimodal Large Language Models (MLLMs) have witnessed rapid advancements in recent years.
Most studies~\citep{qwenvl,Qwen2-VL,Qwen2.5-VL,llava2023,chen2024internvl} developed MLLMs by aligning a visual encoder to a pre-trained LLM through vision-language adaptors (VL-adaptors), making the modality alignment the core of MLLMs development.
Early efforts focused on architectural designs to enhance alignment, exploring various forms of VL-adaptors and visual encoders.
From the perspective of VL-adaptor, three mainstream types have been widely studied: cross-attention modules~\citep{qwenvl,dai2023instructblip,zhang2023internlm}, parallel visual experts~\citep{wang2024cogvlm,dong2024internlm} inspired by LoRA~\citep{hu2022lora}, and simple linear projection layers~\citep{llava2023,li2024llava,Qwen2-VL,Qwen2.5-VL,chen2024expanding}.
Among these, linear projection layers have demonstrated strong effectiveness~\citep{laurenccon2024matters} and are now predominantly adopted in SOTA MLLMs.
Meanwhile, CLIP-based visual encoders have been found to possess intrinsic limitations in multimodal perception~\citep{tong2024eyes,jiang2025devils}, prompting the research of applying hybrid vision towers~\citep{tong2024eyes,lu2024deepseek} and scaling up the vision backbones~\citep{chen2024internvl,chen2024far,chen2024expanding}.
Beyond the exploration of architecture of MLLMs, recent studies have also advanced modality alignment from a data perspective.
Works such as LLaVA~\citep{llava2023,liu2024improved,liu2024llavanext,li2024llava}, Qwen-VL~\citep{qwenvl,Qwen2-VL,Qwen2.5-VL}, and InternVL~\citep{chen2024internvl,chen2024far,chen2024expanding} have significantly scaled up both the volume and diversity of training data.
For instance, LLaVA's training data grew from 753K samples to 9.36M in LLaVA-OV \citep{li2024llava}, while the data diversity broadened from general images to include math reasoning, document and video understanding, substantially improving VL alignment and overall performance on multimodal benchmarks.

\subsection{Multimodal Large Language Models Reasoning}
Since the advent of MLLMs, enhancing their complex multimodal reasoning capabilities has drawn increasing research attention.
Early efforts~\citep{gao2023g,shi2024math,peng2024multimath,xiao2024learning} focused on distilling CoT trajectories from proprietary models like GPT-4V~\citep{achiam2023gpt} and Gemini~\citep{team2023gemini} to inject reasoning abilities into open-source MLLMs.
Although these methods can achieve improvement on targeted benchmarks, they lack generalizability to OOD domains.
Motivated by OpenAI-o1's~\citep{jaech2024openai} test-time scaling, many works~\citep{xu2024llava,xiang2024atomthink,yao2024mulberry,luo2025ursa} explored the implementation of it in multimodal reasoning domain. Approaches like LLaVA-CoT~\citep{xu2024llava}, AtomThink~\citep{xiang2024atomthink}, and URSA~\citep{luo2025ursa} implemented o1-style reasoning by enforcing step-wise outputs and leveraging process-level rewards to evaluate intermediate steps. Recently, the remarkable success of DeepSeek-R1~\citep{guo2025deepseek} in improving LLM reasoning through large-scale RLVR has motivated researchers to transfer similar approaches into the multimodal domain.
Vision-RFT~\citep{liu2025visual} and R1-V~\citep{chen2025r1v} applied RLVR to object detection and counting tasks, significantly improving the image understanding capabilities of MLLMs.
MM-Eureka~\citep{meng2025mm}, VLAA-Thinker~\citep{chen2025sft} and MMR1~\citep{mmr1math2025} extended RLVR on math reasoning tasks without cold-start, achieving substantial improvements in multimodal reasoning. In addition to standard RLVR, R1-VL~\citep{zhang2025r1} and SophiaVL~\citep{fan2025sophiavl} incorporated additional rewards to further supervise the thinking process.
Vision-R1 employed a pipeline that begins with a long CoT cold-start phase and subsequently conducts large-scale RL, leading to superior performance on several multimodal math benchmarks.
Although prior works have made remarkable progress in enhancing the multimodal reasoning abilities of MLLMs, they overlooked the multimodal perception capabilities of MLLMs, which are essential for complex multimodal reasoning and remains difficult to optimize under existing RLVR method due to sparse rewards.

\section{Preliminaries}
This section formulates the multimodal reasoning task (Section~\ref{sec.preliminaries.problem_formulation}) and introduces key concepts of the RLVR algorithm~(Section~\ref{sec.preliminaries.rlvr}) employed in this work.

\subsection{Problem Formulation}\label{sec.preliminaries.problem_formulation}
In this work, we investigate the multimodal reasoning task in the context of MLLMs.
Let $\mathcal{D} = (x_1, x_2, ..., x_N)$ be a multimodal reasoning dataset, where each data sample $x_i = (V, Q, a)$ comprises visual input $V$ (e.g., image), a textual query $Q$, and the corresponding ground-truth answer $a$.
The multimodal reasoning task is defined as follows: given a data sample $x_i \in \mathcal{D}$ as input, the MLLM is required to generate a textual token sequence $y$ that aims to reach the ground-truth answer $a$.

\subsection{Reinforcement Learning with Verifiable Rewards (RLVR)}\label{sec.preliminaries.rlvr}
Reinforcement Learning with Verifiable Rewards (RLVR) is an RL variation that eliminates the dependency on external reward models by using ground-truth answers for reward assignment,
which both mitigates challenging reward hacking issues~\citep{denison2024sycophancy} and substantially reduces computational overhead.
Existing methods~\citep{guo2025deepseek,meng2025mm,huang2025vision} typically apply RLVR with two main components, including reward functions and GRPO algorithm.

\textbf{Reward Functions:} The reward functions consist of two components:
\newline
(1). \textit{Format Reward} ($r_f$) encourages MLLMs to generate in a structured ``think-then-answer'' format, with the reasoning process enclosed in \texttt{<think>} tags and the answer enclosed in \texttt{<answer>} tags.
\newline
(2). \textit{Accuracy Reward} ($r_a$) drives the reasoning optimization in RLVR training by evaluating the correctness of predicted answer. Existing works~\citep{openr1,huang2025vision} mostly adopt a symbolic system to judge the equivalence of ground-truth $a$ and answer in MLLMs' response $y$.

Since format reward $r_f$ only enforces structured output, while accuracy reward $r_a$ plays a central role in enhancing the multimodal reasoning capabilities of MLLMs, we refer to RLVR with a following reward function as \textbf{accuracy-only RLVR}:
\begin{equation}\label{eq.rlvr_reward_funcs}
    r(y_i, a) = \alpha \cdot r_f(y_i) + \beta \cdot r_a(y_i, a)
\end{equation}
where $\alpha, \beta$ are coefficients that control the impact of these two rewards.

\textbf{Group Relative Policy Optimization (GRPO)}~\citep{shao2024deepseekmath} is a variant of Proximal Policy Optimization (PPO)~\citep{schulman2017proximal}, which eliminates the need for a critic by estimating baseline rewards from groups of rollouts, thereby reducing computational overhead while maintaining performance. For each data sample $x \in \mathcal{D}$, GRPO first samples a group of rollouts $Y = (y_1, y_2, ..., y_G)$ from the policy model $\pi_\theta$, then computes advantage $\hat{A}_i$ by normalizing rewards across these rollouts:
\begin{equation}\label{equation.grpo_advantage}
    \hat{A}_{i} = \frac{r(y_i, a) - \mathrm{mean}\{r(y_1, a), r(y_2, a), ..., r(y_G, a)\}}{\mathrm{std}\{r(y_1, a), r(y_2, a), ..., r(y_G, a)\}}
\end{equation}
After obtaining the advantages, GRPO optimizes the policy model $\pi_\theta$ by maximizing the objective:
\begin{equation}\label{equation.grpo_objective}
\begin{aligned}
    &\mathcal{J(\theta)} = \mathbb{E}_{x \in \mathcal{D}, \{y_i\}_{i=1}^G \sim \pi_{\theta_{\mathrm{old}}}} \\
    & \left[\frac{1}{G}\sum_{i=1}^G\frac{1}{|y_i|}\sum_{t=1}^{|y_i|}\left\{\mathrm{min}\left[s_t(x, y_i)\hat{A}_{i}, \mathrm{clip}\left(s_t(x, y_i), 1-\varepsilon, 1+\varepsilon\right)\hat{A}_i\right] - \delta\cdot\mathrm{KL}(\pi_{\theta}\Vert \pi_{ref})\right\}\right]
\end{aligned}
\end{equation}
where $s_t(x, y_i) = \frac{\pi_\theta(y_{i, t}|x, y_{i, <t})}{\pi_{\theta_\mathrm{old}}(y_{i, t}|x, y_{i, <t})}$, $\varepsilon$ is the clipping hyper-parameter, $\delta$ is the coefficient that controls the impact of Kullback-Leibler (KL) divergence, $\pi_\theta$ is the policy model and $\pi_{ref}$ is the fixed reference model that is usually initialized from the initial policy.
\section{Methods}

\subsection{Accuracy-only RLVR Fails to Improve Multimodal Perception in MLLMs}\label{sec.methods.observations}
Multimodal reasoning capabilities can be naturally decomposed into multimodal perception and logical reasoning capabilities~\citep{amizadeh2020neuro,zhou2024proreason}.
Although RLVR has been proven effective in enhancing logical reasoning, many failure cases similar to Figure~\ref{fig.introduction} reveal its limited impact on enhancing multimodal perception. To further validate this observation, we first train Qwen2-VL-7B-IT~\citep{Qwen2-VL} and Qwen2.5-VL-7B-IT~\citep{Qwen2.5-VL} on Geometry3K~\citep{lu2021inter} dataset using accuracy-only RLVR, and then conduct investigations by analyzing CoT trajectories on MathVista~\citep{lu2024mathvista} and MathVerse~\citep{zhang2024mathverse}.

Our investigation yields the following results (Further details are provided in Appendix~\ref{app.sec.analysis_detail}):
\newline
(1). For Qwen2-VL-7B-IT, we analyze 50 and 25 incorrect cases from MathVista and MathVerse, respectively, and find that 72\% and 68\% of these failures are caused by multimodal perception errors. For Qwen2.5-VL-7B-IT, the corresponding proportions are 78\% and 76\%, respectively. These results highlight that multimodal perception remains a major bottleneck for RLVR-trained MLLMs, which limits their further advancement in multimodal reasoning.
\newline
(2). We conduct exact binomial variation of McNemar's test~\citep{McNemar_1947,Edwards_1948} on 50 multimodal problems randomly sampled from MathVista. For Qwen2-VL-7B-IT, the numbers of discordant cases related to multimodal perception are 1 and 5, respectively. For Qwen2.5-VL-7B-IT, the numbers are 2 and 4, respectively. As a result, the exact binomial test yields $p$-values of $0.22$ and $0.69$, both far above the $0.05$ significance level, indicating that the multimodal perception abilities of the accuracy-only RLVR trained MLLMs do not significantly differ from those of the base model.

\subsection{{\methodname}}\label{sec.methods.perception_r1}

\begin{figure}[t]
    \centering
    \includegraphics[width=\linewidth]{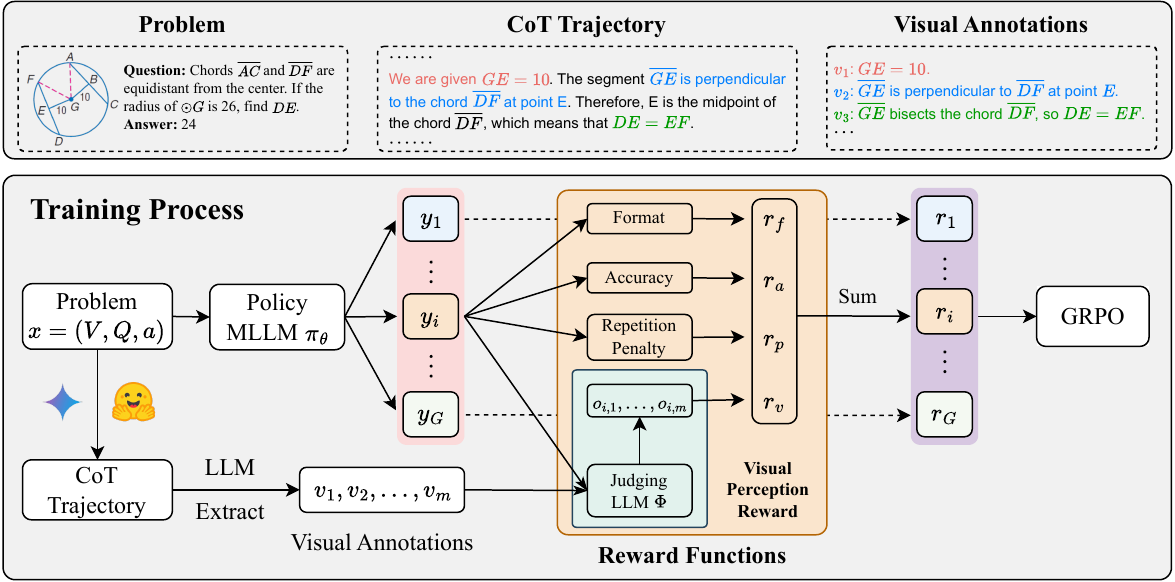}
    \caption{Overview of training pipeline of the proposed {\methodname}. In addition to the accuracy and format rewards, we introduce a novel visual perception reward that explicitly guides MLLMs toward improving their multimodal perception capabilities.}
    \label{fig.methods.method}
\end{figure}

We attribute this limitation to the reward sparsity of accuracy-only RLVR, as answer correctness does not guarantee accurate multimodal perception, (e.g., as illustrated in Figure~\ref{fig.introduction}), making it difficult for accuracy-only RLVR to effectively optimize the multimodal perception capabilities of MLLMs.
To tackle this issue, we propose {\methodname}, which introduces a novel and effective visual perception reward into RLVR, explicitly guiding MLLMs toward improving their multimodal perception capabilities, thereby effectively enhancing their overall multimodal reasoning performance.

Since directly introducing a multimodal reward model may introduce additional reward hacking issues, we largely adhere to the RLVR paradigm in designing the visual perception reward.
As shown in Figure~\ref{fig.methods.method}, we first collect CoT trajectories that contain accurate visual information and then extract visual annotations from them. These visual annotations serve as references for assigning visual perception reward, analogous to the use of ground-truth answers in computing accuracy reward. Subsequently, a judging LLM is used to assess the consistency between visual annotations and the MLLM generated responses, thereby assisting in the assignment of the visual perception reward. Finally, we aggregate all rewards and apply GRPO to optimize the policy model.

\subsubsection{Curation of Visual Annotations}
Visual images often encode rich and complex information that is difficult to convey fully through text. Since our ultimate objective is to enhance the multimodal reasoning capabilities of MLLMs rather than to generate faithful image captions, we focus on guiding MLLMs to concentrate on visual content pertinent to problem solving, such as identifying $GE = 10$ rather than being influenced by superficial cues like line color in Figure~\ref{fig.methods.method}.

To obtain such visual information, we employ a SOTA proprietary MLLM to generate CoT trajectories on multimodal reasoning dataset $\mathcal{D}$, treating the visual information embedded within these trajectories as accurate and highly relevant to problem-solving. Notably, these CoT trajectories can also be obtained from existing open-source multimodal SFT datasets. We then further prompt a strong text-only LLM to extract this embedded visual information from each CoT trajectory into a sequence of visual annotations $\mathcal{V} = (v_1, v_2, ..., v_m)$, where each $v_i$ represents a textual atomic visual annotation of the image that is critical for problem-solving (e.g., $GE = 10$, $\overline{GE} \perp \overline{DF}$ in Figure~\ref{fig.methods.method}), and $m$ denotes the total number of visual annotations within the trajectory. These visual annotations $\mathcal{V}$ will serve as ground-truth references for evaluating whether the policy model accurately perceives visual content during RLVR training, analogous to the role of ground-truth answers in the accuracy reward.

\subsubsection{Visual Perception Reward}
During RLVR training, we need to evaluate the consistency between the visual annotations $\mathcal{V}$ and the visual description embedded in the responses generated by the policy model $\pi_\theta$.
Since symbolic systems struggle to capture the complex semantics of natural language, we address this limitation by introducing a judging LLM $\Phi$ to assess whether each atomic visual annotation $v_i$ is accurately reflected in the responses generated by policy model, thereby extending the source of reward signals.

Formally, given a data sample $x \in \mathcal{D}$ and its corresponding visual annotations $\mathcal{V} = (v_1, v_2, ..., v_m)$, we first sample a response $y_i$ from the policy model $\pi_\theta$, then employ a judging LLM $\Phi$ to assess whether each atom annotation $v_j$ is presented in $y_i$. Consequently, this process results in a judgment sequence $\mathcal{J} = (o_{i,1}, o_{i,2}, ..., o_{i, m})$, where $o_{i, j} \in \{0, 1\}$ indicates whether $v_j$ is accurately reflected in $y_i$ or not. Obtaining $\mathcal{J}$, we can compute the visual perception reward $r_v$ for $y_i$:
\begin{equation}\label{eq.visual_perception_reward}
    r_v(y_i, \mathcal{V}) = \frac{\mathrm{sum}\{o_{i,1}, o_{i,2}, ..., o_{i, m}\}}{|o_{i,1}, o_{i,2}, ..., o_{i, m}|}, \mathrm{where}\ \  o_{i, j} = \Phi(y_i, v_j) \in \{0, 1\}, v_j \in \mathcal{V}
\end{equation}
Accordingly, our visual-enhanced reward function is defined as follows:
\begin{equation}\label{equation.our_rewards}
    r(y_i, a, \mathcal{V}) = \alpha \cdot r_f(y_i) + \beta \cdot r_a(y_i, a) + \gamma \cdot r_v(y_i, \mathcal{V}) + r_p(y_i)
\end{equation}
where $r_f$ and $r_a$ are format reward and accuracy reward explained in Section \ref{sec.preliminaries.rlvr}, $\gamma$ is the coefficient that controls the impact of visual perception reward, $r_p$ is the repetition penalty reward that discourage repetitive behavior during MLLMs' generation. The introduction of $r_p$ is motivated by our observation that directly incorporating $r_v$ will result in increased repetition in the generated responses, which in turn impairs the model's multimodal reasoning capabilities. Following prior works~\citep{yeo2025demystifying,openr1}, we implement $r_p$ using a simple $N$-gram repetition penalty.

During RLVR training, we replace the reward function $r(y_i, a)$ in Eq.\ref{eq.rlvr_reward_funcs} by our visual-enhanced reward $r(y_i, a, \mathcal{V})$, and train the MLLM to maximize the GRPO objective exhibited in Eq.\ref{equation.grpo_objective}.

\section{Experiments}

\definecolor{backblue}{RGB}{210, 230, 250}
\newcommand{\highb}{\cellcolor{backblue}}
\definecolor{backred}{RGB}{255, 223, 223}
\newcommand{\highr}{\cellcolor{backred}}
\definecolor{backgreen}{RGB}{220,244,229}
\newcommand{\highg}{\cellcolor{backgreen}}

\definecolor{textblue}{HTML}{80B7FF}
\definecolor{textred}{HTML}{FF8080}

\subsection{Experiment Settings}\label{sec.experiments.settings}
\textbf{Training Dataset.} We adopt Geometry3K~\citep{lu2021inter} dataset as our training data, which originally contains 2,101 samples for training. To obtain the visual annotations, we employ Gemini-2.5-Pro~\citep{team2023gemini} to generate CoT trajectories on the training data and retain those with correct answers. We then use Qwen2.5-32B-IT~\citep{qwen2.5} to extract visual annotations from the retained CoT trajectories. This process results in a total of 1,442 data samples with associated visual annotations. Model training settings can be found in Appendix~\ref{app.sec.experimental_settings}.

\textbf{Benchmarks and Evalution Settings.} For comprehensive evaluation, we evaluate {\methodname} on a variety of challenging multimodal benchmarks, covering both math and general domains. The math benchmarks include MathVista~\citep{lu2024mathvista}, MathVerse~\citep{zhang2024mathverse}, MathVision~\citep{wang2024measuring} and WeMath~\citep{qiao2024we}. The general benchmarks comprise MMMU~\citep{yue2024mmmu}, MMMU-Pro~\citep{yue2024mmmupro}, MMStar~\citep{chen2024we} and EMMA~\citep{hao2025can}. During inference, we use vLLM~\citep{kwon2023efficient} for efficiency and apply greedy decoding with a temperature of $0.0$.

\textbf{Baselines.} We compare our method against several powerful MLLMs: \textbf{(1) Proprietary MLLMs:} GPT-4o~\citep{hurst2024gpt}, OpenAI-o1~\citep{jaech2024openai}, Claude-3.7-Sonnet~\citep{claude-3.5}, \textbf{(2) Open-Source General MLLMs:} Qwen2-VL-7B-IT~\citep{Qwen2-VL}, Qwen2.5-VL-7B-IT, Qwen2.5-VL-72B-IT~\citep{Qwen2.5-VL}, InternVL2.5-8B~\citep{chen2024expanding}, \textbf{(3) Open-Source Reasoning MLLMs:} URSA-7B~\citep{luo2025ursa}, R1-OneVision~\citep{yang2025r1}, R1-VL~\citep{zhang2025r1}, OpenVLThinker~\citep{deng2025openvlthinker}, VLAA-Thinker~\citep{chen2025sft}, SophiaVL-R1-7B~\citep{fan2025sophiavl}, MM-Eureka~\citep{meng2025mm}, Vision-R1~\citep{huang2025vision}.


\subsection{Main Results}\label{sec.experiments.main_results}

\begin{table}[t]
\centering
\caption{Performance comparison between {\methodname} and baselines on 8 benchmarks. The best and second-best results of Open-Source Reasoning MLLMs are highlighted in {\color{textred}red} and {\color{textblue}blue}. $^\dag$~R1-VL-7B and Vision-R1-7B both trained on WeMath and MathVision, their results are omitted.}
\label{tab.main_results}
\resizebox{1.0\textwidth}{!}{
\begin{tabular}{ll|cccccccc}
\toprule
\multirow{3}{*}{\textbf{Model}} & \multirow{3}{*}{\textbf{\#Data}} & \multicolumn{4}{c}{\textbf{Math Benchmarks}} & \multicolumn{4}{c}{\textbf{General Benchmarks}} \\
\cmidrule(lr){3-6} \cmidrule(lr){7-10}
& & \textbf{MathVista} & \textbf{MathVerse} & \textbf{MathVision} & \textbf{WeMath} & \textbf{MMMU} & \textbf{MMMU-Pro} & \textbf{MMStar} & \textbf{EMMA} \\
& & testmini & testmini & test & testmini & val & overall & val & full \\

\midrule
\multicolumn{10}{c}{ \textit{Proprietary MLLMs} } \\
\midrule
GPT-4o & / & 63.8 & 50.2 & 30.4 & 68.8 & 69.1 & 51.9 & - & 32.7 \\
Claude-3.7-Sonnet & / & 66.8 & 52.0 & 41.3 & 72.6 & 71.0 & 51.5 & 65.1 & 35.1 \\
OpenAI-o1 & / & 73.9 & 57.0 & 60.3 & - & 78.2 & 62.4 & - & 45.7 \\
\midrule
\multicolumn{10}{c}{ \textit{Open Source General MLLMs} } \\
\midrule
Qwen2-VL-7B-IT & / & 58.6 & 31.1 & 16.7 & 42.3 & 46.9 & 29.6 & 56.0 & 24.5 \\
Qwen2.5-VL-7B-IT & / & 68.1 & 47.4 & 25.1 & 61.4 & 55.2 & 37.0 & 63.1 & 24.9 \\
Qwen2.5-VL-72B-IT & / & 75.8 & 55.8 & 37.9 & 71.9 & 70.2 & 49.5 & 70.8 & 38.2 \\
InternVL2.5-8B & / & 64.4 & 39.5 & 19.7 & 53.5 & 56.0 & 34.3 & 62.8 & - \\
\midrule
\multicolumn{10}{c}{ \textit{Open-Source Reasoning MLLMs} } \\
\midrule
URSA-7B & 3.06M & 59.8 & 45.7 & - & - & - & - & - & - \\
R1-VL-7B & 260K & 62.7 & 40.8 & -$^\dag$ & -$^\dag$ & 52.3 & 29.4 & 56.7 & 23.5 \\
R1-OneVision-7B & 155K & 65.0 & 46.5 & 21.9 & 61.9 & 52.9 & 33.8 & 58.9 & 23.6 \\
OpenVLThinker-7B & 25K & 71.3 & 47.4 & 24.3 & \highb{66.3} & \highb{58.4} & 37.8 & 63.8 & 27.0 \\
VLAA-Thinker-7B & 25K & 70.7 & 51.2 & 26.7 & \highb{66.3} & 54.7 & 37.2 & 62.7 & 26.6 \\
SophiaVL-R1-7B & 130K & 70.6 & 49.0 & 26.6 & 64.8 & 56.7 & \highb{38.8} & 63.1 & 27.4 \\
MM-Eureka-7B & 15K & 72.5 & 51.9 & \highb{27.6} & 65.6 & 58.0 & 38.3 & \highb{64.2} & \highb{28.1} \\
Vision-R1-7B & 200K & \highb{73.1} & \highb{52.4} & -$^\dag$ & -$^\dag$ & 55.2 & 37.6 & 62.6 & \highr{\textbf{28.2}} \\
\midrule
\textbf{Perception-R1-7B} & \highr{\textbf{1.4K}} & \highr{\textbf{74.2}} & \highr{\textbf{54.3}} & \highr{\textbf{28.6}} & \highr{\textbf{72.0}} & \highr{\textbf{60.8}} & \highr{\textbf{42.4}} & \highr{\textbf{64.5}} & 27.5 \\
\bottomrule
\end{tabular}
}
\end{table}

We present the performance comparison between our {\methodname} and existing powerful methods across 8 mainstream multimodal benchmarks in Table \ref{tab.main_results}. The performance of applying our method on Qwen2-VL-7B-IT is presented in Appendix~\ref{app.sec.generalize_qwen2}. We summarize our findings as follows:

\textbf{{\methodname} achieves the best performance on most of the benchmarks.} As demonstrated in the table, despite being trained on a small dataset of only 1,442 samples, our {\methodname} still achieves remarkable performance across all benchmarks, outperforming previous powerful methods on all benchmarks except EMMA. We also conduct statistical significance testing using a one-sample t-test, finding that the average improvement is significant with $p < 0.01$ compared to Vision-R1-7B and MM-Eureka-7B. This result provides strong evidence for the superior performance of our proposed {\methodname}. It also underscores the critical role of multimodal perception in enabling effective multimodal reasoning, suggesting that accuracy-only RLVR requires further adaptation when applied to the multimodal reasoning domain. Although {\methodname} is trained on a mere 1,442 math geometry problems, it still achieves the best performance across several general benchmarks. It not only highlights {\methodname}'s superior robustness and generalizability but also demonstrates the critical role of multimodal perception in multimodal reasoning and the rationality of our motivation.

\textbf{The multimodal perception capabilities of {\methodname} show tangible improvements.}
In addition to the overall performance of {\methodname} on benchmarks presented in Table~\ref{tab.main_results}, we provide further evidence for the significant improvement of {\methodname} in multimodal perception capabilities from the following two aspects: (1). We present the performance of {\methodname} and representative baselines on the Vision-Only subsets of the MathVerse and MMMU-Pro benchmarks in Table~\ref{app.tab.vision_only_results}. These subsets exclusively accept images as input, thereby posing a more rigorous challenge to the multimodal perception capabilities of MLLMs. As shown in the table, our {\methodname} still achieves the best performance and outperforms baselines by a large margin, which strongly validates the superior multimodal perception capabilities of {\methodname}.
(2). Similar to statistical test in Section~\ref{sec.methods.observations}, we also conduct McNemar's test on {\methodname}. We investigate the same 50 problems as presented in Section~\ref{sec.methods.observations} and find that the numbers of discordant cases for multimodal perception are 2 and 10, respectively. As a result, the exact binomial variation of McNemar’s test~\citep{McNemar_1947} yields exact $p$ value of $0.04$, below the $0.05$ significance threshold, indicating that the multimodal perception capabilities of {\methodname} is substantially improved compared to the original MLLM.

\textbf{{\methodname} effectively enhances multimodal reasoning capabilities of MLLMs in a highly data-efficient manner.} Although existing methods such as MM-Eureka~\citep{meng2025mm} and Vision-R1~\citep{huang2025vision} have demonstrated strong data efficiency in enhancing the multimodal reasoning capabilities of MLLMs compared to prior SOTA SFT and PRM approach~\citep{luo2025ursa}, our {\methodname} achieves even better performance using over $100\times$ less data than Vision-R1 and $10\times$ less data than MM-Eureka, demonstrating its exceptional data efficiency in developing reasoning MLLMs. This finding suggests that data efficiency can be substantially improved by incorporating richer reward signals from data beyond the final answer, as demonstrated by our proposed visual perception rewards. We believe {\methodname} will achieve further enhanced performance when more high-quality and high-diversity training data is incorporated into its training process in the future.

\subsection{Ablation Study}
\begin{table}[t]
\centering
\caption{Component \& approach ablation studies of {\methodname}. The best result is marked in {\color{textred}{red}}.}
\label{tab.ablation_study}
\resizebox{1.0\textwidth}{!}{
\begin{tabular}{l|cccccccc}
\toprule
\multirow{3}{*}{\textbf{Model}} & \multicolumn{4}{c}{\textbf{Math Benchmarks}} & \multicolumn{4}{c}{\textbf{General Benchmarks}} \\
\cmidrule(lr){2-5} \cmidrule(lr){6-9}
& \textbf{MathVista} & \textbf{MathVerse} & \textbf{MathVision} & \textbf{WeMath} & \textbf{MMMU} & \textbf{MMMU-Pro} & \textbf{MMStar} & \textbf{EMMA} \\
& testmini & testmini & test & testmini & val & overall & val & full \\

\midrule
Qwen2.5-VL-7B-IT & 68.1 & 47.4 & 25.1 & 61.4 & 55.2 & 37.0 & 60.2 & 24.9 \\
+ GRPO & 73.3 & 51.3 & 26.6 & 69.5 & 58.0 & 38.2 & 63.1 & 24.9 \\
\textbf{Perception-R1-7B} & \highr{\textbf{74.2}} & \highr{\textbf{54.3}} & \highr{\textbf{28.6}} & \highr{\textbf{72.0}} & \highr{\textbf{60.8}} & \highr{\textbf{42.4}} & \highr{\textbf{64.5}} & 27.5 \\
\midrule
\multicolumn{9}{c}{ \textit{Component Ablation} } \\
\midrule
w/o Visual Perception Reward & 73.6 & 53.0 & 27.6 & 70.4 & 57.2 & 40.1 & 63.5 & 27.9 \\
w/o Repetition Penalty & 73.6 & 52.6 & 26.9 & 68.5 & 59.1 & 40.6 & 63.6 & 27.6 \\
\midrule
\multicolumn{9}{c}{ \textit{Approach Ablation} } \\
\midrule
Qwen2.5-VL-7B-IT + SFT & 67.3 & 39.1 & 21.3 & 49.1 & 52.8 & 35.2 & 59.6 & \highr{\textbf{28.3}} \\
Qwen2.5-VL-32B-IT as RM & 73.2 & 54.1 & 26.8 & 66.3 & 58.9 & 40.6 & 61.7 & 26.6 \\
\bottomrule
\end{tabular}
}
\end{table}

In this section, we conduct ablation studies from two perspectives: (1) evaluating the effectiveness of each component of {\methodname}, i.e., the visual perception reward and the repetition penalty; and (2) comparing {\methodname} with alternative approaches, including directly using an MLLM as the reward model and employing supervised fine-tuning to train the base model.

We present the results of ablation studies in Table \ref{tab.ablation_study}. As shown in the table, firstly, the accuracy across all benchmarks declines when either the visual perception reward or the repetition penalty is removed, demonstrating the effectiveness and necessity of both components in our {\methodname}. Secondly, all ablations incorporating visual perception reward outperform others that are trained with accuracy-only RLVR on the ``Vision Only'' (VO) subset of MathVerse, further indicating that our proposed visual perception reward enhances the multimodal perception capabilities of MLLMs. Thirdly, directly employing a powerful MLLM (Qwen2.5-VL-32B-IT) as the reward model does not yield better performance than our {\methodname}. We attribute this to reward hacking (See Appendix~\ref{app.sec.reward_model}), which underscores the importance of constructing verifiable visual annotations. To demonstrate the effectiveness of our overall RL training pipeline, we also conduct SFT experiment on the base model using the same 1,442 CoT trajectories distilled from Gemini-2.5-Pro~\citep{team2023gemini}. From Table \ref{tab.ablation_study}, it is observed that the SFT model yields inferior performance, with results on most benchmarks falling short of those of the base model after training. This phenomenon highlights the superior generalization ability and data efficiency of {\methodname} compared to SFT method.

\subsection{Further Analysis of Visual Perception Reward}\label{sec.experiments.visual_reward}

\begin{figure}[t]
  \centering
  \begin{subfigure}[b]{0.333\textwidth}
    \centering
    \includegraphics[width=\linewidth]{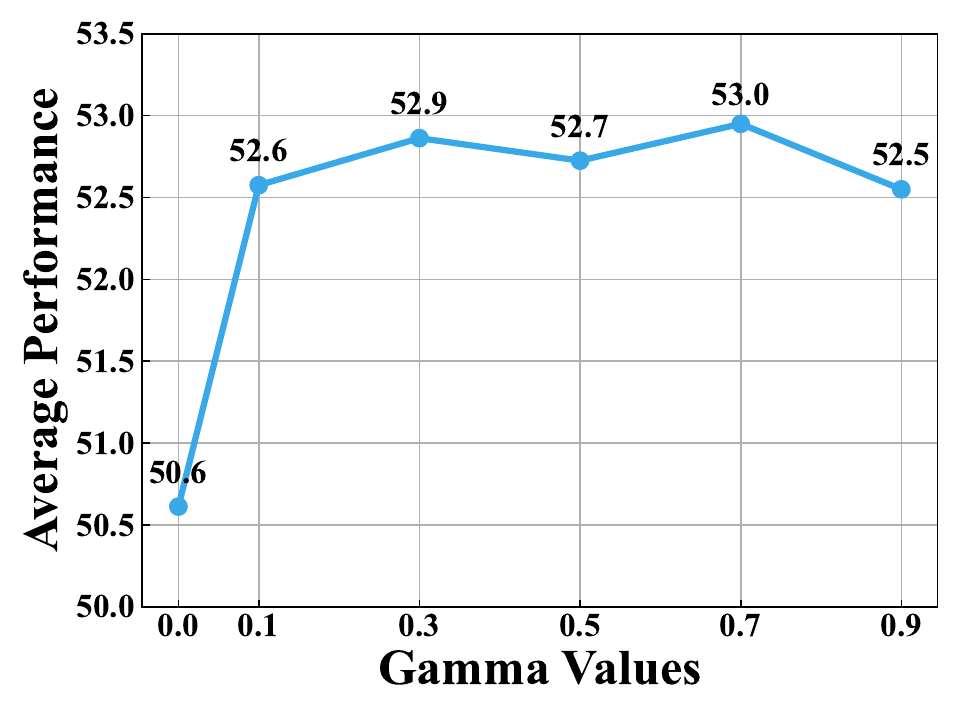}
    \caption{}
    \label{fig.experiments.visual_reward.gamma}
  \end{subfigure}\hfill
  \begin{subfigure}[b]{0.337\textwidth}
    \centering
    \includegraphics[width=\linewidth]{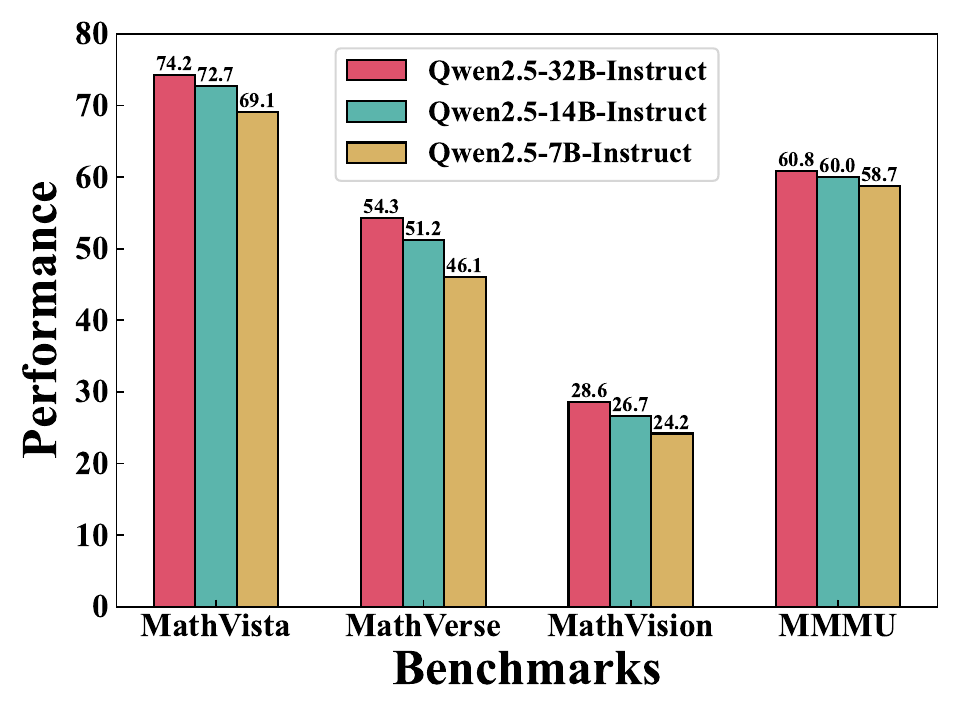}
    \caption{}
    \label{fig.experiments.visual_reward.accuracy}
  \end{subfigure}\hfill
  \begin{subfigure}[b]{0.33\textwidth}
    \centering
    \includegraphics[width=\linewidth]{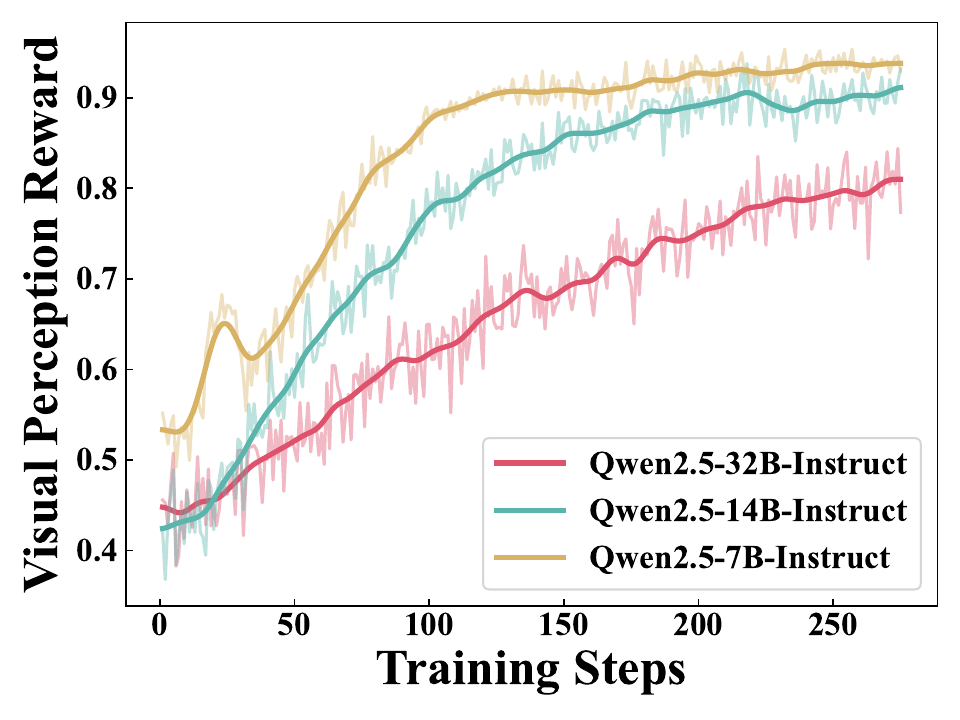}
    \caption{}
    \label{fig.experiments.visual_reward.reward}
  \end{subfigure}

  \caption{(a). Average performance across all benchmarks with varying $\gamma$ values. (b). Comparison of performance across benchmarks when using different judging LLMs. (c). Dynamics of visual perception reward during training when using different judging LLMs.}
  \label{fig.experiments.visual_perception_reward}
\end{figure}

To further explore the dynamics of the visual perception reward, we conduct experiments by varying the coefficient $\gamma$ in Eq.~\ref{equation.our_rewards} and evaluating the impact of different judging LLMs.

To study the impact of coefficient $\gamma$, we train a series of models with $\gamma \in \{0, 0.1, 0.3, 0.5, 0.7, 0.9\}$, and present their average performance across all benchmarks in Figure~\ref{fig.experiments.visual_reward.gamma}. We observe that models trained with different values of $\gamma$ achieve comparable performance across all benchmarks, while all significantly outperform the model that does not incorporate the visual perception reward. The result suggests that only a small amount of visual optimization signal is sufficient to effectively incentivize the multimodal perception and reasoning capabilities of MLLMs, and increasing the value of $\gamma$ does not lead to significantly better results. We attribute this to GRPO, which normalizes the advantages across responses that receive different visual perception rewards when other rewards are identical.

Given that the judging LLM plays a central role in assigning visual perception rewards, we investigate how its capability affects the performance of the resulting MLLM. Specifically, we employ Qwen2.5~\citep{qwen2.5} models of varying sizes as judging LLMs to train Qwen2.5-VL-7B-IT with $\gamma$ fixed at $0.7$. The performance of the resulting models across benchmarks is presented in Figure~\ref{fig.experiments.visual_reward.accuracy}. As the capabilities of the judging LLMs decrease, the performance of the resulting MLLM consistently deteriorates, with the model trained using the 7B judging LLM even underperforming the original MLLM on MathVerse (46.1\% vs. 47.4\%) and MathVision (24.2\% vs. 25.1\%). According to the training dynamics of visual perception reward presented in Figure~\ref{fig.experiments.visual_reward.reward}, the reward increases rapidly and saturates early when using weak judging LLMs, implying the presence of severe reward hacking issues that misguide the resulting MLLM away from accurate problem solving.

\section{Conclusion}
In this paper, we first conduct McNemar’s test on accuracy-only RLVR-trained MLLMs and find no statistically significant improvement in their multimodal perception capabilities compared to their original counterparts, which consequently limits their further advancement in multimodal reasoning.
To address this limitation, we propose {\methodname}, which introduces a novel visual perception reward in addition to the standard accuracy reward, explicitly encouraging accurate visual perception during RLVR training.
Specifically, we first collect textual visual annotations from CoT trajectories as references, and then assign visual perception reward by evaluating the consistency between these annotations and MLLM-generated response using a judging LLM.
Extensive experiments demonstrate the effectiveness of {\methodname}, achieving the best performance compared to multiple baselines on most multimodal math and general benchmarks using only 1,442 training samples.

\section*{Reproducibility Statement}
To ensure the reproducibility of our work, we made efforts in three key areas: providing a clear methodological description, detailing core implementation configurations, and releasing the source code, datasets, and model checkpoints as open-source resources. In Section~\ref{sec.methods.perception_r1}, we present a comprehensive description of the implementation of the visual perception reward and {\methodname}. The prompts used to obtain visual annotations $\mathcal{V}$ and to judge the consistency between the policy model’s response and $\mathcal{V}$ are provided in Appendix~\ref{app.sec.prompts}. Detailed training configurations are listed in Appendix~\ref{app.sec.experimental_settings}. To further support full reproducibility, we include our dataset and training code for {\methodname} in the supplementary materials, and we will release the dataset, source code, and model checkpoint to the community upon publication.

\section*{Acknowledgments}
This research was supported by the National Science and Technology Major Project (No.2023ZD0121102), the National Natural Science Foundation of China (No.U23A20319, No.62477044), Anhui Provincial Natural Science Foundation (No.2308085QF229), the support of the Young Elite Scientists Sponsorship Program by CAST (No.2024QNRC001), and the Fundamental Research Funds for the Central Universities (No.WK2150110038).

\bibliography{refs}

@article{guo2025deepseek,
  title={Deepseek-r1: Incentivizing reasoning capability in llms via reinforcement learning},
  author={Guo, Daya and Yang, Dejian and Zhang, Haowei and Song, Junxiao and Zhang, Ruoyu and Xu, Runxin and Zhu, Qihao and Ma, Shirong and Wang, Peiyi and Bi, Xiao and others},
  journal={arXiv preprint arXiv:2501.12948},
  year={2025}
}

@article{jaech2024openai,
  title={Openai o1 system card},
  author={Jaech, Aaron and Kalai, Adam and Lerer, Adam and Richardson, Adam and El-Kishky, Ahmed and Low, Aiden and Helyar, Alec and Madry, Aleksander and Beutel, Alex and Carney, Alex and others},
  journal={arXiv preprint arXiv:2412.16720},
  year={2024}
}

@article{fan2025sophiavl,
  title={SophiaVL-R1: Reinforcing MLLMs Reasoning with Thinking Reward},
  author={Fan, Kaixuan and Feng, Kaituo and Lyu, Haoming and Zhou, Dongzhan and Yue, Xiangyu},
  journal={arXiv preprint arXiv:2505.17018},
  year={2025}
}

@article{meng2025mm,
  title={MM-Eureka: Exploring Visual Aha Moment with Rule-based Large-scale Reinforcement Learning},
  author={Meng, Fanqing and Du, Lingxiao and Liu, Zongkai and Zhou, Zhixiang and Lu, Quanfeng and Fu, Daocheng and Shi, Botian and Wang, Wenhai and He, Junjun and Zhang, Kaipeng and others},
  journal={arXiv preprint arXiv:2503.07365},
  year={2025}
}

@article{peng2025lmm,
  title={Lmm-r1: Empowering 3b lmms with strong reasoning abilities through two-stage rule-based rl},
  author={Peng, Yingzhe and Zhang, Gongrui and Zhang, Miaosen and You, Zhiyuan and Liu, Jie and Zhu, Qipeng and Yang, Kai and Xu, Xingzhong and Geng, Xin and Yang, Xu},
  journal={arXiv preprint arXiv:2503.07536},
  year={2025}
}

@article{yang2025r1,
  title={R1-onevision: Advancing generalized multimodal reasoning through cross-modal formalization},
  author={Yang, Yi and He, Xiaoxuan and Pan, Hongkun and Jiang, Xiyan and Deng, Yan and Yang, Xingtao and Lu, Haoyu and Yin, Dacheng and Rao, Fengyun and Zhu, Minfeng and others},
  journal={arXiv preprint arXiv:2503.10615},
  year={2025}
}

@article{huang2025vision,
  title={Vision-r1: Incentivizing reasoning capability in multimodal large language models},
  author={Huang, Wenxuan and Jia, Bohan and Zhai, Zijie and Cao, Shaosheng and Ye, Zheyu and Zhao, Fei and Hu, Yao and Lin, Shaohui},
  journal={arXiv preprint arXiv:2503.06749},
  year={2025}
}

@article{deng2025openvlthinker,
  title={Openvlthinker: An early exploration to complex vision-language reasoning via iterative self-improvement},
  author={Deng, Yihe and Bansal, Hritik and Yin, Fan and Peng, Nanyun and Wang, Wei and Chang, Kai-Wei},
  journal={arXiv preprint arXiv:2503.17352},
  year={2025}
}

@inproceedings{lu2024mathvista,
  author    = {Lu, Pan and Bansal, Hritik and Xia, Tony and Liu, Jiacheng and Li, Chunyuan and Hajishirzi, Hannaneh and Cheng, Hao and Chang, Kai-Wei and Galley, Michel and Gao, Jianfeng},
  title     = {MathVista: Evaluating Mathematical Reasoning of Foundation Models in Visual Contexts},
  booktitle={International Conference on Learning Representations (ICLR)},
  year      = {2024}
}

@inproceedings{zhang2024mathverse,
  title={Mathverse: Does your multi-modal llm truly see the diagrams in visual math problems?},
  author={Zhang, Renrui and Jiang, Dongzhi and Zhang, Yichi and Lin, Haokun and Guo, Ziyu and Qiu, Pengshuo and Zhou, Aojun and Lu, Pan and Chang, Kai-Wei and Qiao, Yu and others},
  booktitle={European Conference on Computer Vision},
  pages={169--186},
  year={2024},
  organization={Springer}
}

@inproceedings{wang2024measuring,
  title={Measuring Multimodal Mathematical Reasoning with MATH-Vision Dataset},
  author={Ke Wang and Junting Pan and Weikang Shi and Zimu Lu and Houxing Ren and Aojun Zhou and Mingjie Zhan and Hongsheng Li},
  booktitle={The Thirty-eight Conference on Neural Information Processing Systems Datasets and Benchmarks Track},
  year={2024},
  url={https://openreview.net/forum?id=QWTCcxMpPA}
}

@misc{mmr1math2025,
  title={MMR1: Advancing the Frontiers of Multimodal Reasoning},
  author={Sicong Leng* and Jing Wang* and Jiaxi Li* and Hao Zhang* and Zhiqiang Hu and Boqiang Zhang and Hang Zhang and Yuming Jiang and Xin Li and Deli Zhao and Fan Wang and Yu Rong and Aixin Sun† and Shijian Lu†},
  year={2025},
  howpublished={\url{https://github.com/LengSicong/MMR1}}
}

@article{llava2023,
  title={Visual instruction tuning},
  author={Liu, Haotian and Li, Chunyuan and Wu, Qingyang and Lee, Yong Jae},
  journal={Advances in Neural Information Processing Systems},
  volume={36},
  pages={34892--34916},
  year={2023}
}

@article{qwenvl,
  title={Qwen-VL: A Versatile Vision-Language Model for Understanding, Localization, Text Reading, and Beyond},
  author={Bai, Jinze and Bai, Shuai and Yang, Shusheng and Wang, Shijie and Tan, Sinan and Wang, Peng and Lin, Junyang and Zhou, Chang and Zhou, Jingren},
  journal={arXiv preprint arXiv:2308.12966},
  year={2023}
}

@article{dai2023instructblip,
  title={Instructblip: Towards general-purpose vision-language models with instruction tuning},
  author={Dai, Wenliang and Li, Junnan and Li, Dongxu and Tiong, Anthony and Zhao, Junqi and Wang, Weisheng and Li, Boyang and Fung, Pascale N and Hoi, Steven},
  journal={Advances in Neural Information Processing Systems},
  volume={36},
  pages={49250--49267},
  year={2023}
}

@article{zhang2023internlm,
  title={Internlm-xcomposer: A vision-language large model for advanced text-image comprehension and composition},
  author={Zhang, Pan and Dong, Xiaoyi and Wang, Bin and Cao, Yuhang and Xu, Chao and Ouyang, Linke and Zhao, Zhiyuan and Duan, Haodong and Zhang, Songyang and Ding, Shuangrui and others},
  journal={arXiv preprint arXiv:2309.15112},
  year={2023}
}

@article{wang2024cogvlm,
  title={Cogvlm: Visual expert for pretrained language models},
  author={Wang, Weihan and Lv, Qingsong and Yu, Wenmeng and Hong, Wenyi and Qi, Ji and Wang, Yan and Ji, Junhui and Yang, Zhuoyi and Zhao, Lei and XiXuan, Song and others},
  journal={Advances in Neural Information Processing Systems},
  volume={37},
  pages={121475--121499},
  year={2024}
}

@article{dong2024internlm,
  title={Internlm-xcomposer2: Mastering free-form text-image composition and comprehension in vision-language large model},
  author={Dong, Xiaoyi and Zhang, Pan and Zang, Yuhang and Cao, Yuhang and Wang, Bin and Ouyang, Linke and Wei, Xilin and Zhang, Songyang and Duan, Haodong and Cao, Maosong and others},
  journal={arXiv preprint arXiv:2401.16420},
  year={2024}
}

@article{hu2022lora,
  title={Lora: Low-rank adaptation of large language models.},
  author={Hu, Edward J and Shen, Yelong and Wallis, Phillip and Allen-Zhu, Zeyuan and Li, Yuanzhi and Wang, Shean and Wang, Lu and Chen, Weizhu and others},
  journal={ICLR},
  volume={1},
  number={2},
  pages={3},
  year={2022}
}

@inproceedings{liu2024improved,
  title={Improved baselines with visual instruction tuning},
  author={Liu, Haotian and Li, Chunyuan and Li, Yuheng and Lee, Yong Jae},
  booktitle={Proceedings of the IEEE/CVF Conference on Computer Vision and Pattern Recognition},
  pages={26296--26306},
  year={2024}
}

@misc{qwen3_vl,
  title={Qwen3-VL: Sharper Vision, Deeper Thought, Broader Action},
  author={Qwen Team},
  url={https://qwen.ai/blog?id=99f0335c4ad9ff6153e517418d48535ab6d8afef},
  year={2025},
  urldate={2025-09-23},
}

@misc{qwen3_next,
  title={Qwen3-Next: Towards Ultimate Training \& Inference Efficiency},
  author={Qwen Team},
  url={https://qwen.ai/blog?id=4074cca80393150c248e508aa62983f9cb7d27cd},
  year={2025},
  urldate={2025-09-11},
}

@inproceedings{kafle2018dvqa,
  title={Dvqa: Understanding data visualizations via question answering},
  author={Kafle, Kushal and Price, Brian and Cohen, Scott and Kanan, Christopher},
  booktitle={Proceedings of the IEEE conference on computer vision and pattern recognition},
  pages={5648--5656},
  year={2018}
}

@article{lu2021iconqa,
  title={Iconqa: A new benchmark for abstract diagram understanding and visual language reasoning},
  author={Lu, Pan and Qiu, Liang and Chen, Jiaqi and Xia, Tony and Zhao, Yizhou and Zhang, Wei and Yu, Zhou and Liang, Xiaodan and Zhu, Song-Chun},
  journal={arXiv preprint arXiv:2110.13214},
  year={2021}
}

@inproceedings{cao2022augmented,
  title={An augmented benchmark dataset for geometric question answering through dual parallel text encoding},
  author={Cao, Jie and Xiao, Jing},
  booktitle={Proceedings of the 29th international conference on computational linguistics},
  pages={1511--1520},
  year={2022}
}

@misc{liu2024llavanext,
    title={LLaVA-NeXT: Improved reasoning, OCR, and world knowledge},
    url={https://llava-vl.github.io/blog/2024-01-30-llava-next/},
    author={Liu, Haotian and Li, Chunyuan and Li, Yuheng and Li, Bo and Zhang, Yuanhan and Shen, Sheng and Lee, Yong Jae},
    month={January},
    year={2024}
}

@article{li2024llava,
  title={Llava-onevision: Easy visual task transfer},
  author={Li, Bo and Zhang, Yuanhan and Guo, Dong and Zhang, Renrui and Li, Feng and Zhang, Hao and Zhang, Kaichen and Zhang, Peiyuan and Li, Yanwei and Liu, Ziwei and others},
  journal={arXiv preprint arXiv:2408.03326},
  year={2024}
}

@inproceedings{mitra2024compositional,
  title={Compositional chain-of-thought prompting for large multimodal models},
  author={Mitra, Chancharik and Huang, Brandon and Darrell, Trevor and Herzig, Roei},
  booktitle={Proceedings of the IEEE/CVF Conference on Computer Vision and Pattern Recognition},
  pages={14420--14431},
  year={2024}
}

@article{zhang2023multimodal,
  title={Multimodal Chain-of-Thought Reasoning in Language Models},
  author={Zhang, Zhuosheng and Zhang, Aston and Li, Mu and Karypis, George and Smola, Alex and others},
  journal={Transactions on Machine Learning Research},
  year={2023}
}

@article{liu2024socraticlm,
  title={SocraticLM: Exploring socratic personalized teaching with large language models},
  author={Liu, Jiayu and Huang, Zhenya and Xiao, Tong and Sha, Jing and Wu, Jinze and Liu, Qi and Wang, Shijin and Chen, Enhong},
  journal={Advances in Neural Information Processing Systems},
  volume={37},
  pages={85693--85721},
  year={2024}
}

@article{xue2024decompose,
  title={Decompose, analyze and rethink: Solving intricate problems with human-like reasoning cycle},
  author={Xue, Shangzi and Huang, Zhenya and Liu, Jiayu and Lin, Xin and Ning, Yuting and Jin, Binbin and Li, Xin and Liu, Qi},
  journal={Advances in Neural Information Processing Systems},
  volume={37},
  pages={357--385},
  year={2024}
}

@article{xiao2024learning,
  title={Learning to solve geometry problems via simulating human dual-reasoning process},
  author={Xiao, Tong and Liu, Jiayu and Huang, Zhenya and Wu, Jinze and Sha, Jing and Wang, Shijin and Chen, Enhong},
  journal={arXiv preprint arXiv:2405.06232},
  year={2024}
}

@inproceedings{jiang2025devils,
  title={Devils in middle layers of large vision-language models: Interpreting, detecting and mitigating object hallucinations via attention lens},
  author={Jiang, Zhangqi and Chen, Junkai and Zhu, Beier and Luo, Tingjin and Shen, Yankun and Yang, Xu},
  booktitle={Proceedings of the Computer Vision and Pattern Recognition Conference},
  pages={25004--25014},
  year={2025}
}

@article{Qwen2.5-VL,
  title={Qwen2.5-VL Technical Report},
  author={Bai, Shuai and Chen, Keqin and Liu, Xuejing and Wang, Jialin and Ge, Wenbin and Song, Sibo and Dang, Kai and Wang, Peng and Wang, Shijie and Tang, Jun and Zhong, Humen and Zhu, Yuanzhi and Yang, Mingkun and Li, Zhaohai and Wan, Jianqiang and Wang, Pengfei and Ding, Wei and Fu, Zheren and Xu, Yiheng and Ye, Jiabo and Zhang, Xi and Xie, Tianbao and Cheng, Zesen and Zhang, Hang and Yang, Zhibo and Xu, Haiyang and Lin, Junyang},
  journal={arXiv preprint arXiv:2502.13923},
  year={2025}
}

@inproceedings{kwon2023efficient,
  title={Efficient Memory Management for Large Language Model Serving with PagedAttention},
  author={Woosuk Kwon and Zhuohan Li and Siyuan Zhuang and Ying Sheng and Lianmin Zheng and Cody Hao Yu and Joseph E. Gonzalez and Hao Zhang and Ion Stoica},
  booktitle={Proceedings of the ACM SIGOPS 29th Symposium on Operating Systems Principles},
  year={2023}
}

@article{zheng2023ddcot,
  title={Ddcot: Duty-distinct chain-of-thought prompting for multimodal reasoning in language models},
  author={Zheng, Ge and Yang, Bin and Tang, Jiajin and Zhou, Hong-Yu and Yang, Sibei},
  journal={Advances in Neural Information Processing Systems},
  volume={36},
  pages={5168--5191},
  year={2023}
}

@article{zhang2024cocot,
  title={Cocot: Contrastive chain-of-thought prompting for large multimodal models with multiple image inputs},
  author={Zhang, Daoan and Yang, Junming and Lyu, Hanjia and Jin, Zijian and Yao, Yuan and Chen, Mingkai and Luo, Jiebo},
  journal={arXiv preprint arXiv:2401.02582},
  year={2024}
}

@article{Qwen2-VL,
  title={Qwen2-VL: Enhancing Vision-Language Model's Perception of the World at Any Resolution},
  author={Wang, Peng and Bai, Shuai and Tan, Sinan and Wang, Shijie and Fan, Zhihao and Bai, Jinze and Chen, Keqin and Liu, Xuejing and Wang, Jialin and Ge, Wenbin and Fan, Yang and Dang, Kai and Du, Mengfei and Ren, Xuancheng and Men, Rui and Liu, Dayiheng and Zhou, Chang and Zhou, Jingren and Lin, Junyang},
  journal={arXiv preprint arXiv:2409.12191},
  year={2024}
}

@article{team2025kimi,
  title={Kimi k1. 5: Scaling reinforcement learning with llms},
  author={Team, Kimi and Du, Angang and Gao, Bofei and Xing, Bowei and Jiang, Changjiu and Chen, Cheng and Li, Cheng and Xiao, Chenjun and Du, Chenzhuang and Liao, Chonghua and others},
  journal={arXiv preprint arXiv:2501.12599},
  year={2025}
}

@article{chen2024expanding,
  title={Expanding performance boundaries of open-source multimodal models with model, data, and test-time scaling},
  author={Chen, Zhe and Wang, Weiyun and Cao, Yue and Liu, Yangzhou and Gao, Zhangwei and Cui, Erfei and Zhu, Jinguo and Ye, Shenglong and Tian, Hao and Liu, Zhaoyang and others},
  journal={arXiv preprint arXiv:2412.05271},
  year={2024}
}

@article{laurenccon2024matters,
  title={What matters when building vision-language models?},
  author={Lauren{\c{c}}on, Hugo and Tronchon, L{\'e}o and Cord, Matthieu and Sanh, Victor},
  journal={Advances in Neural Information Processing Systems},
  volume={37},
  pages={87874--87907},
  year={2024}
}

@inproceedings{chen2024internvl,
  title={Internvl: Scaling up vision foundation models and aligning for generic visual-linguistic tasks},
  author={Chen, Zhe and Wu, Jiannan and Wang, Wenhai and Su, Weijie and Chen, Guo and Xing, Sen and Zhong, Muyan and Zhang, Qinglong and Zhu, Xizhou and Lu, Lewei and others},
  booktitle={Proceedings of the IEEE/CVF conference on computer vision and pattern recognition},
  pages={24185--24198},
  year={2024}
}

@inproceedings{zhang2024mavis,
  title={MAVIS: Mathematical Visual Instruction Tuning with an Automatic Data Engine},
  author={Zhang, Renrui and Wei, Xinyu and Jiang, Dongzhi and Guo, Ziyu and Zhang, Yichi and Tong, Chengzhuo and Liu, Jiaming and Zhou, Aojun and Zhang, Shanghang and Gao, Peng and others},
  booktitle={The Thirteenth International Conference on Learning Representations},
  year={2024}
}

@inproceedings{tong2024eyes,
  title={Eyes wide shut? exploring the visual shortcomings of multimodal llms},
  author={Tong, Shengbang and Liu, Zhuang and Zhai, Yuexiang and Ma, Yi and LeCun, Yann and Xie, Saining},
  booktitle={Proceedings of the IEEE/CVF Conference on Computer Vision and Pattern Recognition},
  pages={9568--9578},
  year={2024}
}

@article{lu2024deepseek,
  title={Deepseek-vl: towards real-world vision-language understanding},
  author={Lu, Haoyu and Liu, Wen and Zhang, Bo and Wang, Bingxuan and Dong, Kai and Liu, Bo and Sun, Jingxiang and Ren, Tongzheng and Li, Zhuoshu and Yang, Hao and others},
  journal={arXiv preprint arXiv:2403.05525},
  year={2024}
}

@article{chen2024far,
  title={How Far Are We to GPT-4V? Closing the Gap to Commercial Multimodal Models with Open-Source Suites},
  author={Chen, Zhe and Wang, Weiyun and Tian, Hao and Ye, Shenglong and Gao, Zhangwei and Cui, Erfei and Tong, Wenwen and Hu, Kongzhi and Luo, Jiapeng and Ma, Zheng and others},
  journal={arXiv preprint arXiv:2404.16821},
  year={2024}
}

@inproceedings{gao2023g,
  title={G-LLaVA: Solving Geometric Problem with Multi-Modal Large Language Model},
  author={Gao, Jiahui and Pi, Renjie and Zhang, Jipeng and Ye, Jiacheng and Zhong, Wanjun and Wang, Yufei and Hong, Lanqing and Han, Jianhua and Xu, Hang and Li, Zhenguo and others},
  booktitle={The Thirteenth International Conference on Learning Representations},
  year={2024}
}

@article{achiam2023gpt,
  title={Gpt-4 technical report},
  author={Achiam, Josh and Adler, Steven and Agarwal, Sandhini and Ahmad, Lama and Akkaya, Ilge and Aleman, Florencia Leoni and Almeida, Diogo and Altenschmidt, Janko and Altman, Sam and Anadkat, Shyamal and others},
  journal={arXiv preprint arXiv:2303.08774},
  year={2023}
}

@article{team2023gemini,
  title={Gemini: a family of highly capable multimodal models},
  author={Team, Gemini and Anil, Rohan and Borgeaud, Sebastian and Alayrac, Jean-Baptiste and Yu, Jiahui and Soricut, Radu and Schalkwyk, Johan and Dai, Andrew M and Hauth, Anja and Millican, Katie and others},
  journal={arXiv preprint arXiv:2312.11805},
  year={2023}
}

@article{peng2024multimath,
  title={Multimath: Bridging visual and mathematical reasoning for large language models},
  author={Peng, Shuai and Fu, Di and Gao, Liangcai and Zhong, Xiuqin and Fu, Hongguang and Tang, Zhi},
  journal={arXiv preprint arXiv:2409.00147},
  year={2024}
}

@inproceedings{shi2024math,
  title={Math-LLaVA: Bootstrapping Mathematical Reasoning for Multimodal Large Language Models},
  author={Shi, Wenhao and Hu, Zhiqiang and Bin, Yi and Liu, Junhua and Yang, Yang and Ng, See Kiong and Bing, Lidong and Lee, Roy},
  booktitle={Findings of the Association for Computational Linguistics: EMNLP 2024},
  pages={4663--4680},
  year={2024}
}

@article{xiang2024atomthink,
  title={AtomThink: A Slow Thinking Framework for Multimodal Mathematical Reasoning},
  author={Xiang, Kun and Liu, Zhili and Jiang, Zihao and Nie, Yunshuang and Huang, Runhui and Fan, Haoxiang and Li, Hanhui and Huang, Weiran and Zeng, Yihan and Han, Jianhua and others},
  journal={arXiv preprint arXiv:2411.11930},
  year={2024}
}

@article{luo2025ursa,
  title={URSA: Understanding and Verifying Chain-of-thought Reasoning in Multimodal Mathematics},
  author={Luo, Ruilin and Zheng, Zhuofan and Wang, Yifan and Yu, Yiyao and Ni, Xinzhe and Lin, Zicheng and Zeng, Jin and Yang, Yujiu},
  journal={arXiv preprint arXiv:2501.04686},
  year={2025}
}

@article{xu2024llava,
  title={Llava-o1: Let vision language models reason step-by-step},
  author={Xu, Guowei and Jin, Peng and Hao, Li and Song, Yibing and Sun, Lichao and Yuan, Li},
  journal={arXiv preprint arXiv:2411.10440},
  year={2024}
}

@article{yao2024mulberry,
  title={Mulberry: Empowering mllm with o1-like reasoning and reflection via collective monte carlo tree search},
  author={Yao, Huanjin and Huang, Jiaxing and Wu, Wenhao and Zhang, Jingyi and Wang, Yibo and Liu, Shunyu and Wang, Yingjie and Song, Yuxin and Feng, Haocheng and Shen, Li and others},
  journal={arXiv preprint arXiv:2412.18319},
  year={2024}
}

@article{liu2025visual,
  title={Visual-rft: Visual reinforcement fine-tuning},
  author={Liu, Ziyu and Sun, Zeyi and Zang, Yuhang and Dong, Xiaoyi and Cao, Yuhang and Duan, Haodong and Lin, Dahua and Wang, Jiaqi},
  journal={arXiv preprint arXiv:2503.01785},
  year={2025}
}

@misc{chen2025r1v,
  author       = {Chen, Liang and Li, Lei and Zhao, Haozhe and Song, Yifan and Vinci},
  title        = {R1-V: Reinforcing Super Generalization Ability in Vision-Language Models with Less Than \$3},
  howpublished = {\url{https://github.com/Deep-Agent/R1-V}},
  year         = {2025}
}

@article{denison2024sycophancy,
  title={Sycophancy to subterfuge: Investigating reward-tampering in large language models},
  author={Denison, Carson and MacDiarmid, Monte and Barez, Fazl and Duvenaud, David and Kravec, Shauna and Marks, Samuel and Schiefer, Nicholas and Soklaski, Ryan and Tamkin, Alex and Kaplan, Jared and others},
  journal={arXiv preprint arXiv:2406.10162},
  year={2024}
}

@misc{openr1,
    title = {Open R1: A fully open reproduction of DeepSeek-R1},
    url = {https://github.com/huggingface/open-r1},
    author = {Hugging Face},
    month = {January},
    year = {2025}
}

@article{shao2024deepseekmath,
  title={Deepseekmath: Pushing the limits of mathematical reasoning in open language models},
  author={Shao, Zhihong and Wang, Peiyi and Zhu, Qihao and Xu, Runxin and Song, Junxiao and Bi, Xiao and Zhang, Haowei and Zhang, Mingchuan and Li, YK and Wu, Y and others},
  journal={arXiv preprint arXiv:2402.03300},
  year={2024}
}

@article{schulman2017proximal,
  title={Proximal policy optimization algorithms},
  author={Schulman, John and Wolski, Filip and Dhariwal, Prafulla and Radford, Alec and Klimov, Oleg},
  journal={arXiv preprint arXiv:1707.06347},
  year={2017}
}

@article{qwen2.5,
    title   = {Qwen2.5 Technical Report}, 
    author  = {An Yang and Baosong Yang and Beichen Zhang and Binyuan Hui and Bo Zheng and Bowen Yu and Chengyuan Li and Dayiheng Liu and Fei Huang and Haoran Wei and Huan Lin and Jian Yang and Jianhong Tu and Jianwei Zhang and Jianxin Yang and Jiaxi Yang and Jingren Zhou and Junyang Lin and Kai Dang and Keming Lu and Keqin Bao and Kexin Yang and Le Yu and Mei Li and Mingfeng Xue and Pei Zhang and Qin Zhu and Rui Men and Runji Lin and Tianhao Li and Tingyu Xia and Xingzhang Ren and Xuancheng Ren and Yang Fan and Yang Su and Yichang Zhang and Yu Wan and Yuqiong Liu and Zeyu Cui and Zhenru Zhang and Zihan Qiu},
    journal = {arXiv preprint arXiv:2412.15115},
    year    = {2024}
}

@inproceedings{amizadeh2020neuro,
  title={Neuro-symbolic visual reasoning: Disentangling},
  author={Amizadeh, Saeed and Palangi, Hamid and Polozov, Alex and Huang, Yichen and Koishida, Kazuhito},
  booktitle={International Conference on Machine Learning},
  pages={279--290},
  year={2020},
  organization={Pmlr}
}

@article{zhou2024proreason,
  title={ProReason: Multi-Modal Proactive Reasoning with Decoupled Eyesight and Wisdom},
  author={Zhou, Jingqi and Wang, Sheng and Dong, Jingwei and Li, Lei and Gao, Jiahui and Kong, Lingpeng and Wu, Chuan},
  journal={arXiv preprint arXiv:2410.14138},
  year={2024}
}

@inproceedings{lu2021inter,
  title={Inter-GPS: Interpretable Geometry Problem Solving with Formal Language and Symbolic Reasoning},
  author={Lu, Pan and Gong, Ran and Jiang, Shibiao and Qiu, Liang and Huang, Siyuan and Liang, Xiaodan and Zhu, Song-chun},
  booktitle={Proceedings of the 59th Annual Meeting of the Association for Computational Linguistics and the 11th International Joint Conference on Natural Language Processing (Volume 1: Long Papers)},
  pages={6774--6786},
  year={2021}
}

@article{McNemar_1947,
    title={Note on the Sampling Error of the Difference Between Correlated Proportions or Percentages},
    volume={12},
    DOI={10.1007/BF02295996},
    number={2},
    journal={Psychometrika},
    author={McNemar, Quinn},
    year={1947},
    pages={153–157}
}

@article{Edwards_1948,
    title={Note on the “Correction for Continuity” in Testing the Significance of the Difference between Correlated Proportions},
    volume={13},
    DOI={10.1007/BF02289261},
    number={3},
    journal={Psychometrika},
    author={Edwards, Allen L.},
    year={1948},
    pages={185–187}
}

@article{yeo2025demystifying,
  title={Demystifying Long Chain-of-Thought Reasoning in LLMs},
  author={Yeo, Edward and Tong, Yuxuan and Niu, Morry and Neubig, Graham and Yue, Xiang},
  journal={arXiv preprint arXiv:2502.03373},
  year={2025}
}

@article{qiao2024we,
  title={We-math: Does your large multimodal model achieve human-like mathematical reasoning?},
  author={Qiao, Runqi and Tan, Qiuna and Dong, Guanting and Wu, Minhui and Sun, Chong and Song, Xiaoshuai and GongQue, Zhuoma and Lei, Shanglin and Wei, Zhe and Zhang, Miaoxuan and others},
  journal={arXiv preprint arXiv:2407.01284},
  year={2024}
}

@misc{zheng2025easyr1,
  title        = {EasyR1: An Efficient, Scalable, Multi-Modality RL Training Framework},
  author       = {Yaowei Zheng and Junting Lu and Shenzhi Wang and Zhangchi Feng and Dongdong Kuang and Yuwen Xiong},
  howpublished = {\url{https://github.com/hiyouga/EasyR1}},
  year         = {2025}
}

@article{chen2025sft,
  title={Sft or rl? an early investigation into training r1-like reasoning large vision-language models},
  author={Chen, Hardy and Tu, Haoqin and Wang, Fali and Liu, Hui and Tang, Xianfeng and Du, Xinya and Zhou, Yuyin and Xie, Cihang},
  journal={arXiv preprint arXiv:2504.11468},
  year={2025}
}

@misc{claude-3.5,
  author       = {Anthropic},
  title        = {Claude 3.5 sonnet},
  howpublished = {\url{https://www.anthropic.com/news/claude-3-5-sonnet}},
  year         = {2024}
}

@article{hurst2024gpt,
  title={Gpt-4o system card},
  author={Hurst, Aaron and Lerer, Adam and Goucher, Adam P and Perelman, Adam and Ramesh, Aditya and Clark, Aidan and Ostrow, AJ and Welihinda, Akila and Hayes, Alan and Radford, Alec and others},
  journal={arXiv preprint arXiv:2410.21276},
  year={2024}
}

@inproceedings{yue2024mmmu,
  title={Mmmu: A massive multi-discipline multimodal understanding and reasoning benchmark for expert agi},
  author={Yue, Xiang and Ni, Yuansheng and Zhang, Kai and Zheng, Tianyu and Liu, Ruoqi and Zhang, Ge and Stevens, Samuel and Jiang, Dongfu and Ren, Weiming and Sun, Yuxuan and others},
  booktitle={Proceedings of the IEEE/CVF Conference on Computer Vision and Pattern Recognition},
  pages={9556--9567},
  year={2024}
}

@article{chen2024we,
  title={Are we on the right way for evaluating large vision-language models?},
  author={Chen, Lin and Li, Jinsong and Dong, Xiaoyi and Zhang, Pan and Zang, Yuhang and Chen, Zehui and Duan, Haodong and Wang, Jiaqi and Qiao, Yu and Lin, Dahua and others},
  journal={Advances in Neural Information Processing Systems},
  volume={37},
  pages={27056--27087},
  year={2024}
}

@inproceedings{hao2025can,
  title={Can MLLMs Reason in Multimodality? EMMA: An Enhanced MultiModal ReAsoning Benchmark},
  author={Hao, Yunzhuo and Gu, Jiawei and Wang, Huichen Will and Li, Linjie and Yang, Zhengyuan and Wang, Lijuan and Cheng, Yu},
  booktitle={International Conference on Machine Learning},
  year={2025}
}

@article{yue2024mmmupro,
  title={Mmmu-pro: A more robust multi-discipline multimodal understanding benchmark},
  author={Yue, Xiang and Zheng, Tianyu and Ni, Yuansheng and Wang, Yubo and Zhang, Kai and Tong, Shengbang and Sun, Yuxuan and Yu, Botao and Zhang, Ge and Sun, Huan and others},
  journal={arXiv preprint arXiv:2409.02813},
  year={2024}
}

@article{zhang2025r1,
  title={R1-vl: Learning to reason with multimodal large language models via step-wise group relative policy optimization},
  author={Zhang, Jingyi and Huang, Jiaxing and Yao, Huanjin and Liu, Shunyu and Zhang, Xikun and Lu, Shijian and Tao, Dacheng},
  journal={arXiv preprint arXiv:2503.12937},
  year={2025}
}

@article{yu2025dapo,
  title={Dapo: An open-source llm reinforcement learning system at scale},
  author={Yu, Qiying and Zhang, Zheng and Zhu, Ruofei and Yuan, Yufeng and Zuo, Xiaochen and Yue, Yu and Fan, Tiantian and Liu, Gaohong and Liu, Lingjun and Liu, Xin and others},
  journal={arXiv preprint arXiv:2503.14476},
  year={2025}
}
\bibliographystyle{iclr2026_conference}


\newpage
\appendix
\section{Benchmarks and Baselines}\label{app.benchmarks_baselines}

\subsection{Benchmarks}

\begin{itemize}[leftmargin=3ex,itemsep=1ex,topsep=0ex]
    \item \textbf{MathVista}~\citep{lu2024mathvista} MathVista is a consolidated benchmark for multimodal mathematical reasoning. We evaluate our {\methodname} and all baselines on \texttt{testmini} split of MathVista, which consists of five subtasks: Textbook Question Answering, Visual Question Answering, Geometry Problem Solving, Math Word Problems and Figure Question Answering.
    \item \textbf{MathVerse}~\citep{zhang2024mathverse} MathVerse is a benchmark designed to evaluate the reasoning capabilities of MLLMs under varying proportions of textual and visual information. We evaluate our {\methodname} and all baselines on \texttt{testmini} split of MathVerse, which includes 5 subset: Text Dominant (TD), Text Lite (TL), Vision Intensive (VI), Vision Dominant (VD), and Vision Only (VO). MathVerse covers three subtasks: ``Plane Geometry'' with 2,550 problems, ``Functions'' with 795 problems and ``Solid Geometry'' with 595 problems.
    \item \textbf{MathVision}~\citep{wang2024measuring} MathVision consists of 3,040 high quality mathematical problems with visual contexts sourced from real math competitions. We evaluate our {\methodname} and all baselines on \texttt{test} split of MathVision, which includes five difficulty levels and 16 subtasks.
    \item \textbf{WeMath}~\citep{qiao2024we} WeMath is the benchmark specifically designed to explore the problem-solving principles beyond the end-to-end performance, spanning 67 hierarchical knowledge concepts and 5 layers of knowledge granularity. We evaluate our {\methodname} and all baselines on \texttt{testmini} split of WeMath under the multiple-choice setting.
    \item \textbf{MMMU}~\citep{yue2024mmmu} MMMU is a widely used multi-discipline multimodal benchmark that covers a broad scope of tasks, including Art, Business, Health \& Medicine, Science, Humanities \& Social Science, and Tech \& Engineering, and over subfields, thus can comprehensively assess the multimodal reasoning abilities of a MLLM. We evaluate our {\methodname} and all baselines on \texttt{val} subset of MMMU.
    \item \textbf{MMMU-Pro}~\citep{yue2024mmmupro} MMMU-Pro is a more robust version of MMMU benchmark. MMMU-Pro improves MMMU from following 3 perspectives: (1). Excluding the problem that can be answered by text-only models, (2). augmenting the candidate options of multiple choice problems, making guessing more infeasible, and (3). introducing vision-only input setting where questions are embedded within images.
    \item \textbf{MMStar}~\citep{chen2024we} MMStar comprises 1,500 meticulously curated problems sourced from a diverse range of existing multimodal benchmarks. To guarantee exceptional quality, the selection of problems for MMStar adheres to two core principles: (1). Visual information must be indispensable to solving the problem, and (2). avoiding data leakage.
    \item \textbf{EMMA}~\citep{hao2025can} EMMA is also a multi-discipline multimodal benchmark that covers mathematics, physics, chemistry, and coding. Different from previous multimodal benchmarks, EMMA emphasize the importance of organically reason over and with both text and images, therefore places higher requirements on the multimodal reasoning capabilities of MLLMs.
\end{itemize}

\subsection{Baselines}

\begin{itemize}[leftmargin=3ex,itemsep=1ex,topsep=0ex]
    \item \textbf{URSA-7B}~\citep{luo2025ursa} URSA enhanced the multimodal reasoning capabilities of MLLMs through an SFT approach. It employed a three-part data synthesis strategy to construct a high-quality CoT reasoning dataset for SFT. USRA further incorporated a dual-view trajectory labeling approach, resulting in the DualMath-1.1M dataset, and trained a PRM to achieve test-time scaling.
    \item \textbf{R1-VL-7B}~\citep{zhang2025r1} R1-VL proposed a new online reinforcement learning framework StepGRPO, which enabled MLLMs to self-improve reasoning ability via simple, effective and dense step-wise rewarding. It consisted of two dense reasoning rewards: StepRAR and StepRVR. StepRAR was used to reward the accurate intermediate reasoning steps and StepRVR was used to reward the well-structure of the overall reasoning path.
    \item \textbf{R1-OneVision-7B}~\citep{yang2025r1} R1-OneVision adopted a cold-start then RL training pipeline to enhance the reasoning capabilities of MLLMs. It first addressed the modality gap to construct a high-quality long CoT multimodal dataset for cold-start initialization, then applied accuracy-only RLVR on 10K randomly sampled data to further incentivize MLLM's reasoning abilities.
    \item \textbf{OpenVLThinker-7B}~\citep{deng2025openvlthinker} OpenVLThinker adopted an approach that iteratively leverages SFT on lightweight training data and RL to improve reasoning capabilities of MLLMs. During the training pipeline, OpenVLThinker progressively evolved the data across iterations, retaining more challenging examples for later stages of training.
    \item \textbf{VLAA-Thinker-7B}~\citep{chen2025sft} VLAA-Thinker was developed by directly conducting RL on the VLAA-Thinking-RL-25K dataset using Qwen2.5-VL models. The main contributions of VLAA-Thinker were twofold: (1). VLAA-Thinking-RL-25K dataset was constructed by carefully selecting multimodal data from a variety of existing multimodal datasets. (2). It proposed a mixed reward approach to train MLLM during RL.
    \item \textbf{SophiaVL-R1-7B}~\citep{fan2025sophiavl} SophiaVL-R1 argued that outcome-based rewards alone cannot ensure a high-quality thinking process. To address this, it first trained a thinking reward model to evaluate the reasoning quality of intermediate steps. This reward model was then incorporated into RL to provide an additional thinking reward, guiding the policy model to generate trajectories with more coherent and well-reasoned intermediate steps.
    \item \textbf{MM-Eureka-7B}~\citep{meng2025mm} MM-Eureka constructed the MMK12 dataset, which contained 15,616 high quality multimodal reasoning data, and then directly applied accuracy-only RLVR on this dataset. To stabilize and improve training, MM-Eureka incorporated several techniques during RL, including online data filtering, the removal of KL penalty, and a two-stage training strategy.
    \item \textbf{Vision-R1-7B}~\citep{huang2025vision} Vision-R1 adopted a two-stage pipeline consisting of cold-start initialization followed by reinforcement learning to enhance the multimodal reasoning capabilities of MLLMs. It first employed powerfull MLLMs and DeepSeek-R1~\citep{guo2025deepseek} to fill the modality gap and curate 200K multimodal CoT data for cold-start initialization. In the second stage, Vision-R1 applied accuracy-only RLVR on an additional 10K math problems, incorporating the proposed Progressive Thinking Suppression Training (PTST) technique.
\end{itemize}

\section{Further Results}

\begin{table}[t]
  \centering
  \caption{Confusion matrix of Qwen2-VL-7B-IT evaluated on $\mathcal{D}_e$.}
  \label{app.table.qwen2_confusion_raw}
  \begin{tabular}{l|cc}
    \toprule
     & Correct Answer & Wrong Answer \\
    \midrule
    Correct Perception & 15 & 4 \\
    Wrong Perception & 4 & 27 \\
    \bottomrule
  \end{tabular}
\end{table}

\begin{table}[t]
  \centering
  \caption{Confusion matrix of accuracy-only RLVR trained Qwen2-VL-7B-IT evaluated on $\mathcal{D}_e$.}
  \label{app.table.qwen2_confusion_trained}
  \begin{tabular}{l|cc}
    \toprule
     & Correct Answer & Wrong Answer \\
    \midrule
    Correct Perception & 23 & 0 \\
    Wrong Perception & 5 & 22 \\
    \bottomrule
  \end{tabular}
\end{table}

\begin{table}[t]
  \centering
  \caption{Confusion matrix of Qwen2.5-VL-7B-IT evaluated on $\mathcal{D}_e$.}
  \label{app.table.confusion_raw}
  \begin{tabular}{l|cc}
    \toprule
     & Correct Answer & Wrong Answer \\
    \midrule
    Correct Perception & 19 & 3 \\
    Wrong Perception & 11 & 17 \\
    \bottomrule
  \end{tabular}
\end{table}

\begin{table}[t]
  \centering
  \caption{Confusion matrix of accuracy-only RLVR trained Qwen2.5-VL-7B-IT evaluated on $\mathcal{D}_e$.}
  \label{app.table.confusion_trained}
  \begin{tabular}{l|cc}
    \toprule
     & Correct Answer & Wrong Answer \\
    \midrule
    Correct Perception & 23 & 1 \\
    Wrong Perception & 15 & 11 \\
    \bottomrule
  \end{tabular}
\end{table}

\subsection{Details of Analysis of Accuracy-only RLVR-trained MLLMs}\label{app.sec.analysis_detail}
In this section, we present additional details and results to the analysis in Section~\ref{sec.methods.observations}.

We conduct our investigation from the following two perspectives: 
(1) The proportion of incorrect solving cases attributable to multimodal perception errors, and (2) a comparative analysis of the multimodal perception capabilities between the RLVR-trained MLLMs and their original counterparts. The former helps identify the bottleneck in the multimodal reasoning abilities of MLLMs, while the latter assesses whether their perception capabilities improve after accuracy-only RLVR training.

Specifically, we first train Qwen2-VL-7B-IT~\citep{Qwen2-VL} and Qwen2.5-VL-7B-IT~\citep{Qwen2.5-VL} models on Geometry3K~\citep{lu2021inter} dataset using accuracy-only RLVR. We then manually assess their CoT trajectories on the geometry reasoning subset of MathVista~\citep{lu2024mathvista}, as well as the “Visual Dominant” and “Visual Only” subsets of MathVerse~\citep{zhang2024mathverse}. These problems require both strong multimodal perception and logical reasoning capabilities, making them suitable for identifying potential weaknesses in MLLMs’ reasoning performance. For each multimodal problem, we consider an MLLM to have made a perception error if its CoT trajectory contains an inaccurate visual description that is essential for reaching the correct final answer. All annotations are conducted by three well-trained annotators (all with at least a bachelor’s degree). The template for human annotation is shown in Figure~\ref{app.fig.human_annotation}.

Let $\mathcal{D}_e$ denote the set of 50 problems randomly sampled from MathVista in Section~\ref{sec.methods.observations}. We provide additional results for Qwen2-VL-7B-IT in Tables~\ref{app.table.qwen2_confusion_raw} and~\ref{app.table.qwen2_confusion_trained}, and for Qwen2.5-VL-7B-IT in Tables~\ref{app.table.confusion_raw} and~\ref{app.table.confusion_trained}, where each problem is categorized based on the correctness of the model’s final answer and visual perception.

Taking Qwen2.5-VL-7B-IT as an example, from the table, we observe that although the RLVR-trained model shows a significant improvement in problem-solving accuracy (from 30 to 38, i.e., 60\% to 76\%), its visual perception accuracy improves only marginally (from 22 to 24, i.e., 44\% to 48\%). Moreover, the proportion of problems with incorrect visual perception among those solved correctly even increases slightly (from 11/30 to 15/38, i.e., 36.7\% to 39.4\%), which also indicates that the multimodal perception capabilities of the RLVR-trained model have not been effectively improved.

\begin{figure}[t]
\centering
\begin{tcolorbox}
\textbf{[Question ID]}

\textbf{[Image]}
\newline

\textbf{Question:} Chords \(\overline{AC}\) and \(\overline{DF}\) are equidistant from the center. If the radius of \(\odot G\) is 26, find \(DE\).

\textbf{Answer:} 24

\textbf{Response:} \textless think\textgreater  To find the length of \(DE\), we need to use the property that chords equidistant from the center of a circle are congruent. This means that \(AC = DF\).

$\cdots\cdots$

Since \(AC\) and \(DF\) are equidistant and congruent in length, \(DF = 48\). Applying the same logic to chord \(\overline{DF}\), we use the right triangle \(\triangle DGE\):
\(DG = \sqrt{RG^2 - GE^2} = \sqrt{26^2 - 10^2} = 24\).
So, \(DF = 2 \times DE\). Since \(DF = 48\), we have: \(DE = \frac{DF}{2} = \frac{48}{2} = 24\)
\textless /think\textgreater 
\textless answer\textgreater \(\boxed{24}\)\textless /answer\textgreater 

\textbf{Please check whether there exist any multimodal perception error in Response that is critical for problem solving.}

\textbf{Judgment:} [Yes / No]

\end{tcolorbox}
\caption{The template of human evaluation for multimodal perception.}
\label{app.fig.human_annotation}
\end{figure}


\subsection{Manual Examination of Generated Visual Annotations}\label{app.sec.visual_examination}
The correctness of the generated visual annotations is crucial for the effective application of the visual perception reward. To ensure their quality, we manually examined 100 randomly selected retrained samples to identify potential perception errors. Our analysis revealed that only 4 out of 100 samples contained minor perceptual inaccuracies, corresponding to a 96\% perceptual accuracy in the visual annotations, which ensures an effective training of our {\methodname}.

\begin{table}[t]
\centering
\caption{Classification results of extracted visual annotations from Geometry3K and GeoQA+.}
\label{app.table.annotation_classification}
\resizebox{1.0\textwidth}{!}{
\begin{tabular}{l|cccc}
    \toprule
    Dataset & All annotations & Visual Information & Reasoning Information & Text Information \\
    \midrule
    Geometry3K & 99 & 81 (82\%) & 10 (10\%) & 8 (8\%) \\
    GeoQA+ & 114 & 29 (25\%) & 41 (36\%) & 44 (39\%) \\
    \bottomrule
\end{tabular}
}
\end{table}

The extracted visual annotations may contains implicitly reasoning process from the collected CoT trajectories, leading to implicit reasoning distillation. Although including implicit reasoning processes can also enhance the MLLM's logical reasoning capabilities through RL, this is inconsistent with our original intention of improving the MLLM’s perceptual capabilities via the visual perception reward. To demonstrate that the performance gains of our {\methodname} indeed stem from improved perception, we rigorously categorize the extracted visual annotations into ``Visual Information'', ``Reasoning Information'', and ``Text Information''. Here, ``Visual Information'' refers to information that can only be obtained from the image; ``Reasoning Information'' refers to information that does not directly exist in the image or problem text, but can be obtained through reasoning; and ``Text Information'' refers to information that can be obtained directly from the problem text. For comparison, we also conducted the same examination on the GeoQA+ dataset. We randomly examined 30 data instances sampled from the Geometry3K~\citep{lu2021inter} and GeoQA+~\citep{cao2022augmented} datasets, and we present the results in Table~\ref{app.table.annotation_classification}.

It is worth noting that any annotation that requires even mild reasoning to obtain is classified as ``Reasoning Information'' (e.g., Given an image of parallelogram ABCD, if the annotation states that ``AB = CD'', then it is classified as reasoning information). Most ``Reasoning Information'' are such short statements rather than reasoning chains, minimizing the possibility of implicit reasoning distillation. Even with such a rigorous principle, the proportion of visual information in Geometry3K is still dominant (82\%). In comparison, the proportion is only 25\% for the GeoQA+ dataset. This is the core reason why we chose the Geometry3K dataset as our training data, as it provides better visual perception for our framework and isolates it from the influence of implicit reasoning.

\subsection{Generalization Analysis of {\methodname}}\label{app.sec.generalize_qwen2}
\subsubsection{Generalize to Qwen2-VL Model}
We apply our visual perception reward enhanced RLVR to train Qwen2-VL-7B-IT~\citep{Qwen2-VL} to demonstrate its generalizability and robustness. We present the experimental results in Table~\ref{app.tab.qwen2_ablation_study}. Here, we compare against R1-VL~\citep{zhang2025r1}, as it is also trained from Qwen2-VL-7B-IT. From the table, we observe that {\methodname}-Qwen2 achieves the best performance on most benchmarks except MathVision~\citep{wang2024measuring}, demonstrating the effectiveness and generalizability of our method. Notably, similar to the full {\methodname}, {\methodname}-Qwen2 achieves a substantial improvement on the “Vision Only” subset of MathVerse~\citep{zhang2024mathverse} (39.2\% vs. 30.1\%), further validating the effectiveness of the proposed visual perception reward in enhancing the multimodal perception capabilities of MLLMs. We attribute the sub-optimal performance of {\methodname}-Qwen2 on MathVision to the limited diversity of the training dataset and believe this can be addressed by scaling up both the quantity and diversity of the training data.
\begin{table}[t]
\centering
\caption{Experimental results of applying our method to Qwen2-VL-7B-IT. The best result is highlighted in {\color{textred}{red}}. $^\dag$~R1-VL-7B used WeMath and MathVision for training, their results on these benchmarks are omitted.}
\label{app.tab.qwen2_ablation_study}
\resizebox{1.0\textwidth}{!}{
\begin{tabular}{ll|cccccccc}
\toprule
\multirow{3}{*}{\textbf{Model}} & \multirow{3}{*}{\textbf{\#Data}} & \multicolumn{4}{c}{\textbf{Math Benchmarks}} & \multicolumn{4}{c}{\textbf{General Benchmarks}} \\
\cmidrule(lr){3-6} \cmidrule(lr){7-10}
& & \textbf{MathVista} & \textbf{MathVerse} & \textbf{MathVision} & \textbf{WeMath} & \textbf{MMMU} & \textbf{MMMU-Pro} & \textbf{MMStar} & \textbf{EMMA} \\
& & testmini & testmini & test & testmini & val & overall & val & full \\

\midrule
Qwen2-VL-7B-IT & / & 58.6 & 31.1 & 16.7 & 42.3 & 46.9 & 29.6 & 56.0 & 24.5 \\
+ GRPO & 1.4K & 64.5 & 38.1 & 19.7 & 54.6 & 51.4 & 32.4 & 56.3 & 24.3 \\
\midrule
R1-VL-7B & 10K & 62.7 & 40.8 & -$^\dag$ & -$^\dag$ & 52.3 & 29.4 & 56.7 & 23.5 \\
\midrule
\textbf{Perception-R1-Qwen2-7B} & \highr{\textbf{1.4K}} & \highr{\textbf{64.9}} & \highr{\textbf{42.3}} & \highr{\textbf{20.4}} & \highr{\textbf{60.0}} & \highr{\textbf{53.1}} & \highr{\textbf{35.2}} & \highr{\textbf{56.9}} & \highr{\textbf{25.1}} \\
\bottomrule
\end{tabular}
}
\end{table}

\subsubsection{Generalize to Mulberry Dataset}
To further demonstrate the generalizability of our method to other datasets, we conducted the same training pipeline on the data filtered from mulberry-260k~\citep{yao2024mulberry}, which contains 16.8K data and mainly from IconQA~\citep{lu2021iconqa}, DVQA~\citep{kafle2018dvqa} and does not contain any geometry data. During data collection stage, we employ Qwen3-VL-235B-A22B-Instruct~\citep{qwen3_vl} model to generate reasoning trajectories and employ Qwen3-Next-80B-A3B-Instruct~\citep{qwen3_next} model to extract visual annotations because of their powerful multimodal reasoning and language understanding capabilities. The prompts used in data collection and collection pipeline are same as those in Section~\ref{app.sec.prompts}. We name the model trained on this dataset ``Perception-R1-Mulberry-7B''. The experimental results are present in Table~\ref{app.tab.generalize_to_mulberry}.

\begin{table}[t]
\centering
\caption{Experimental results of applying our method to our filtered mulberry dataset. The best result is highlighted in {\color{textred}{red}}.}
\label{app.tab.generalize_to_mulberry}
\resizebox{1.0\textwidth}{!}{
\begin{tabular}{ll|cccccccc}
\toprule
\multirow{3}{*}{\textbf{Model}} & \multirow{3}{*}{\textbf{\#Data}} & \multicolumn{4}{c}{\textbf{Math Benchmarks}} & \multicolumn{4}{c}{\textbf{General Benchmarks}} \\
\cmidrule(lr){3-6} \cmidrule(lr){7-10}
& & \textbf{MathVista} & \textbf{MathVerse} & \textbf{MathVision} & \textbf{WeMath} & \textbf{MMMU} & \textbf{MMMU-Pro} & \textbf{MMStar} & \textbf{EMMA} \\
& & testmini & testmini & test & testmini & val & overall & val & full \\
\midrule
Qwen2.5-VL-7B-IT & / & 68.1 & 47.4 & 25.1 & 61.4 & 55.2 & 37.0 & 60.2 & 24.9 \\
\midrule
+ GRPO on Geometry3K & 1.4K & 73.3 & 51.3 & 26.6 & 69.5 & 58.0 & 38.2 & 63.1 & 24.9 \\
\textbf{Perception-R1-7B} & 1.4K & \highr{\textbf{74.2}} & \highr{\textbf{54.3}} & \highr{\textbf{28.6}} & \highr{\textbf{72.0}} & \highr{\textbf{60.8}} & \highr{\textbf{42.4}} & \highr{\textbf{64.5}} & \highr{\textbf{27.5}} \\
\midrule
+ GRPO on Mulberry & 16.8K & 72.6 & 46.2 & \highr{\textbf{27.8}} & 66.8 & 52.1 & 42.0 & 62.1 & 26.4 \\
\textbf{Perception-R1-Mulberry-7B} & 16.8K & \highr{\textbf{73.4}} & \highr{\textbf{51.2}} & 27.1 & \highr{\textbf{69.9}} & \highr{\textbf{59.1}} & \highr{\textbf{42.2}} & \highr{\textbf{62.6}} & \highr{\textbf{27.2}} \\
\bottomrule
\end{tabular}
}
\end{table}

From Table~\ref{app.tab.generalize_to_mulberry}, we can observe that Perception-R1-Mulberry-7B still outperforms standard GRPO by 2.1 points on average across all benchmarks, demonstrating the effectiveness of our method. We believe the reason why Perception-R1-7B outperforms Perception-R1-Mulberry-7B is that the collected Mulberry data lacks math reasoning content (especially geometry) and mainly focuses on pure visual perception, which leads to worse performance on math benchmarks.

\subsection{Robustness Analysis of {\methodname}}

\begin{table}[t]
\centering
\caption{Experimental results of randomly flipping judgment results $o_{i, j}$ at different proportions.}
\label{app.tab.noise_robustness}
\resizebox{1.0\textwidth}{!}{
\begin{tabular}{l|ccccccccc}
\toprule
\multirow{3}{*}{\textbf{Random flipping proportion}} & \multicolumn{4}{c}{\textbf{Math Benchmarks}} & \multicolumn{4}{c}{\textbf{General Benchmarks}} & \multirow{3}{*}{\textbf{Average}} \\
\cmidrule(lr){2-5} \cmidrule(lr){6-9}
& \textbf{MathVista} & \textbf{MathVerse} & \textbf{MathVision} & \textbf{WeMath} & \textbf{MMMU} & \textbf{MMMU-Pro} & \textbf{MMStar} & \textbf{EMMA} & \\
& testmini & testmini & test & testmini & val & overall & val & full & \\

\midrule
Qwen2.5-VL-7B-IT + GRPO & 73.3 & 51.3 & 26.6 & 69.5 & 58.0 & 38.2 & 63.1 & 24.9 & 50.6 \\
\midrule
0\% (Perception-R1) & 74.2 & 54.3 & 28.6 & 72.0 & 60.8 & 42.4 & 64.5 & 27.5 & 53.0 \\
10\% & 72.2 & 51.1 & 29.1 & 69.2 & 60.5 & 42.9 & 63.1 & 28.3 & 51.9 \\
20\% & 70.0 & 50.6 & 27.4 & 70.7 & 60.9 & 42.0 & 62.1 & 27.9 & 51.5 \\
\bottomrule
\end{tabular}
}
\end{table}

\begin{table}[t]
\centering
\caption{Experimental results of models trained with different $\gamma$ values.}
\label{app.tab.gamma_robustness}
\resizebox{1.0\textwidth}{!}{
\begin{tabular}{l|ccccccccc}
\toprule
\multirow{3}{*}{\textbf{$\gamma$ values}} & \multicolumn{4}{c}{\textbf{Math Benchmarks}} & \multicolumn{4}{c}{\textbf{General Benchmarks}} & \multirow{3}{*}{\textbf{Average}} \\
\cmidrule(lr){2-5} \cmidrule(lr){6-9}
& \textbf{MathVista} & \textbf{MathVerse} & \textbf{MathVision} & \textbf{WeMath} & \textbf{MMMU} & \textbf{MMMU-Pro} & \textbf{MMStar} & \textbf{EMMA} & \\
& testmini & testmini & test & testmini & val & overall & val & full & \\

\midrule
0.0 (GRPO) & 73.3 & 51.3 & 26.6 & 69.5 & 58.0 & 38.2 & 63.1 & 24.9 & 50.6 \\
\midrule
0.1 & 72.7 & 54.1 & 28.5 & 70.9 & 60.0 & 41.2 & 65.4 & 27.8 & 52.6 \\
0.3 & 73.0 & 54.4 & 29.0 & 71.7 & 60.5 & 42.6 & 63.7 & 28.1 & 52.9 \\
0.5 & 75.5 & 53.0 & 27.6 & 70.5 & 59.1 & 42.9 & 65.5 & 27.4 & 52.7 \\
0.7 (Perception-R1) & 74.2 & 54.3 & 28.6 & 72.0 & 60.8 & 42.4 & 64.5 & 27.5 & 53.0 \\
0.9 & 72.4 & 53.7 & 28.4 & 72.2 & 60.9 & 40.7 & 64.1 & 28.0 & 52.5 \\
\bottomrule
\end{tabular}
}
\end{table}

There are two factors can affect the robustness of {\methodname}: the correctness of visual perception reward and the factor $\gamma$ that controls the influence of visual perception reward to the final reward.
\begin{itemize}[leftmargin=3ex,itemsep=1ex,topsep=0ex]
    \item Regarding the correctness of the visual perception reward, there are two types of factors that can impair it: the correctness of the extracted visual annotations and the correctness of the judgments produced by the judging LLM. We simulate these two types of noise by randomly flipping the judgments (i.e., $o_{i,j}$ in Eq.~\ref{eq.visual_perception_reward}) produced by the Qwen2.5-32B-Instruct model from $1 \rightarrow 0$ or $0 \rightarrow 1$ at a fixed proportion. We conduct experiments with flipping proportions of 10\% and 20\%, and present the results in Table~\ref{app.tab.noise_robustness}. From the table, we can observe that even with 20\% of the visual perception reward corrupted, the model’s average performance still surpasses that of GRPO, showcasing the robustness of our method. Notably, the performance degradation mainly comes from MathVista and MathVerse. This may be because these two benchmarks contain a large number of geometry test cases that are similar to our training data. In general benchmarks including MMMU and MMMU-Pro, the model trained with corrupted annotations still performs on par with Perception-R1-7B, further demonstrating the robustness of our training pipeline.
    \item Regarding the factor $\gamma$, we present in Table~\ref{app.tab.gamma_robustness} the performance of models trained with different $\gamma$ values on each benchmark, as an extension of the average performance shown in Figure~\ref{fig.experiments.visual_reward.gamma}. From the table, we observe that the average performances of models trained with different $\gamma$ values (except 0.0) are very similar, and all of them significantly surpass standard GRPO, demonstrating the robustness and effectiveness of our proposed method.
\end{itemize}

\subsection{Performance on Vision-Only Benchmarks}\label{app.sec.vision_only}
\begin{table}[t]
\centering
\caption{Performance comparisons between Perception-R1 and baselines on vision-only subsets of MathVerse and MMMU-Pro. The best result is highlighted in \color{textred}{red}.}
\label{app.tab.vision_only_results}
\resizebox{0.58\textwidth}{!}{
\begin{tabular}{l|cc}
\toprule
\multirow{2}{*}{\textbf{Model}} & \textbf{MathVerse} & \textbf{MMMU-Pro} \\
& vision-only & vision \\

\midrule
\multicolumn{3}{c}{\textit{Qwen2.5-VL Models}} \\
\midrule
Qwen2.5-VL-7B-IT & 42.2 & 33.8 \\
+ GRPO & 47.1 & 37.1 \\
\midrule
R1-Onevision-7B & 41.9 & 30.7 \\
OpenVLThinker-7B & 39.5 & 35.3 \\
VLAA-Thinker-7B & 45.7 & 34.8 \\
SophiaVL-R1-7B & 43.3 & 37.6 \\
MM-Eureka-7B & 47.6 & 35.2 \\
Vision-R1-7B & 47.0 & 36.0 \\
\midrule
\textbf{Perception-R1-7B} & \highr{\textbf{50.1}} & \highr{\textbf{40.3}} \\
$\Delta$ (Ours - Prev SOTA) & \textbf{+2.5} & \textbf{+2.7} \\
\midrule
\multicolumn{3}{c}{\textit{Qwen2-VL Models}} \\
\midrule
Qwen2-VL-7B-IT & 30.1 & 26.6 \\
+ GRPO & 32.4 & 29.8 \\
\midrule
R1-VL-7B & 36.8 & 23.6 \\
\midrule
\textbf{Perception-R1-Qwen2-7B} & \highr{\textbf{39.2}} & \highr{\textbf{33.7}} \\
$\Delta$ (Ours - Prev SOTA) & \textbf{+2.4} & \textbf{+10.1} \\
\bottomrule
\end{tabular}
}
\end{table}

To further demonstrate the improved perception capabilities of Perception-R1 and Perception-R1-Qwen2 models, we compare their performance with baseline methods on Vision-Only subsets of MathVerse~\citep{zhang2024mathverse} and MMMU-Pro~\citep{yue2024mmmupro} benchmarks in Table~\ref{app.tab.vision_only_results}.

From the table, we can observe that both Perception-R1 and Perception-R1-Qwen2 surpass standard GRPO and previous SOTA method on these two vision-only benchmarks by a substantial margin. Specifically, Perception-R1 achieves an \textbf{average improvement of 2.6}, while Perception-R1-Qwen2 reaches an \textbf{average improvement of 6.3}. These results not only demonstrate that the multimodal perception capabilities of the Perception-R1 model series have been significantly enhanced but also validate the effectiveness of our proposed visual perception reward in boosting the multimodal perception capabilities of MLLMs.

\subsection{Analysis of Using Qwen2.5-VL-32B-IT as Reward Model}\label{app.sec.reward_model}
\begin{figure}[t]
  \centering
  \begin{subfigure}[b]{0.49\textwidth}
    \centering
    \includegraphics[width=\linewidth]{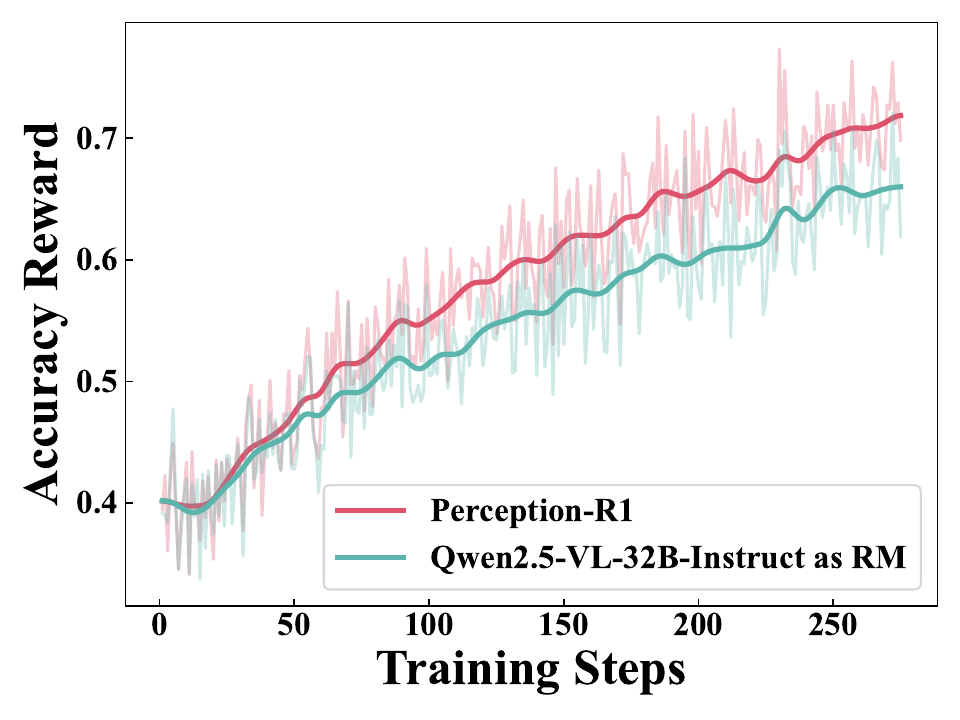}
    \caption{Dynamics of Accuracy Reward.}
    \label{app.fig.experiments.vl_32b_reward_model.accuracy_reward}
  \end{subfigure}\hfill
  \begin{subfigure}[b]{0.49\textwidth}
    \centering
    \includegraphics[width=\linewidth]{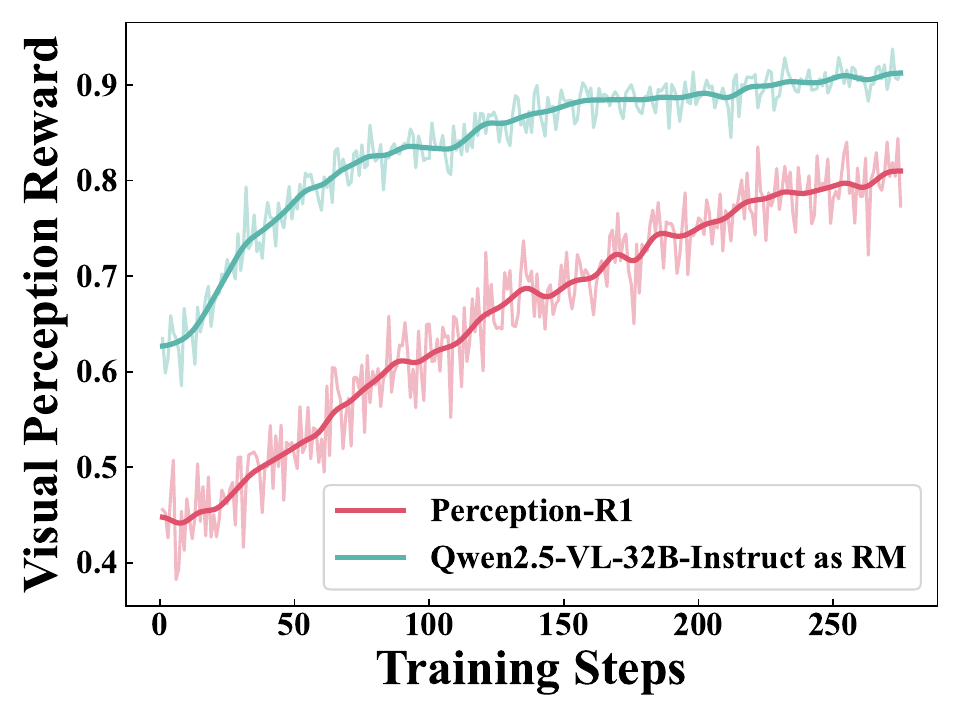}
    \caption{Dynamics of Visual Perception Reward.}
    \label{app.fig.experiments.vl_32b_reward_model.visual_reward}
  \end{subfigure}

  \caption{Comparison of Accuracy and Visual Perception Rewards between {\methodname} and the variant using Qwen2.5-VL-32B-IT as the Reward Model.}
  \label{app.fig.experiments.vl_32b_reward_model}
\end{figure}

We provide the training dynamics of accuracy reward and visual perception reward of {\methodname} and the variant using Qwen2.5-VL-32B-IT as the reward model in Figure~\ref{app.fig.experiments.vl_32b_reward_model}. When using Qwen2.5-VL-32B-IT as the reward model, we provide it with both the image and the response generated by the policy model, and prompt it to output a consistency score in $[0, 1]$, representing the degree of alignment between the image and the response. From Figure~\ref{app.fig.experiments.vl_32b_reward_model}, we observe that the visual perception reward increases rapidly and saturates around 100 training steps. Meanwhile, the accuracy reward becomes consistently lower than that of {\methodname} after the same point, indicating the presence of reward hacking when using Qwen2.5-VL-32B-IT as the reward model. This reward hacking issue undermines the multimodal reasoning performance of the resulting MLLM.

\subsection{Computational Costs Comparison}\label{app.sec.computational_costs}
In this subsection, we compare the computational costs of {\methodname} with representative baseline methods. We categorize the computational costs into \textbf{data preparation cost} and \textbf{training time cost}. For data preparation cost, we estimate it by counting the generated tokens in data curation process using Qwen2.5 Tokenizer. For training time cost, we calculate the total GPU-Hours used to train the model.
We summarize the data preparation costs and training time costs of {\methodname} and representative baseline methods in Table~\ref{app.tab.computational_costs}, with detailed explanations provided below:

\textbf{Data Preparation Cost}:
\begin{itemize}[leftmargin=3ex,itemsep=1ex,topsep=0ex]
    \item \textbf{{\methodname}}: We collected CoT trajectories on 2,101 data samples (before filtering), resulting in a total of 1.01M tokens. For visual annotation extraction, we generated an additional 105K tokens. Thus, the total token cost is 1.1M tokens.
    \item \textbf{Vision-R1} prompted DeepSeek-R1 to produce 200K CoT trajectories. The total number of generated tokens is 134M.
    \item \textbf{MM-Eureka} performed pure RL on 15K self-collected samples without trajectory distillation, resulting in 0 token generation cost.
    \item \textbf{SophiaVL-R1} constructed the large-scale SophiaVL-R1-Thinking-156K dataset to train a thinking reward model for evaluating the thinking quality of the policy model during RL. This dataset was built by collecting CoT trajectories and leveraging powerful MLLM-based judgments, resulting in a total of 39.4M tokens.
    \item \textbf{VLAA-Thinker}: Although VLAA-Thinker did not perform SFT, its RL training dataset (VLAA-Thinking-Dataset) was selected and constructed by analyzing the captions and CoT trajectories generated by GPT-4o and DeepSeek-R1. Here we only count the tokens of RL dataset, which resulting in a total of 29.6M tokens.
    \item \textbf{OpenVLThinker} distilled 25K samples, of which only 3.2K (731K tokens) are publicly available. We estimate the total token count to be about 5.71M.
    \item \textbf{R1-OneVision} heavily relied on GPT-4o to enhance a subset of the LLaVA-OneVision dataset, but the augmented data is hard to separate from the original trajectories, making token cost estimation infeasible. Nonetheless, with 155K samples collected for SFT, its token generation cost likely exceeds 1.1M with high probability.
    \item \textbf{R1-VL} used the off-the-shelf mulberry-260K dataset for SFT, resulting in 0 token cost.
\end{itemize}

\textbf{Training Time Cost}:
Since only Vision-R1 reported its detailed training setup, we can only provide a detailed comparison with it. For other baselines (R1-VL, R1-OneVision and OpenVLThinker), which require large-scale SFT, their training costs likely exceed that of {\methodname} with high probability.
\begin{itemize}[leftmargin=3ex,itemsep=1ex,topsep=0ex]
    \item \textbf{{\methodname}} can be trained in 16 hours using 16 A800 GPUs: 8 for serving the judging LLM and 8 for policy training. Each RL step takes an average of 154.3s, with judgment accounting for 47.5s, which means the 8 serving GPUs are idle 69.2\% of the time and can be used for other API tasks. The total training cost of {\methodname} is about 167.4 A800-Hours. Compared to standard GRPO, the only additional cost comes from judgment, increasing per-step time by 44.5\%. However, due to the effectiveness of visual perception reward, {\methodname} requires significantly less data than other baselines, ultimately resulting in a substantially lower total training cost.
    \item \textbf{Vision-R1} needed 32 H800 GPUs for about 10h SFT and 64 H800 GPUs for about 2 day RL, resulting in a total training cost of ~3,392 H800-hours.
    \item \textbf{MM-Eureka} was trained on 15K distinct samples for 10 epochs, resulting in a total of 150K samples used for policy training. In contrast, {\methodname} was trained on 1.4K distinct samples for 25 epochs, yielding 35K samples in total. Therefore, the training time cost of MM-Eureka is larger than that of {\methodname} with high probability.
    \item \textbf{SophiaVL-R1} was trained on 130K RL samples, which required first training a thinking reward model on 158K SFT samples. During RL, SophiaVL-R1 needed incorporating the reward model into the RL training process. Therefore, the training time cost of SophiaVL-R1 is larger than that of {\methodname} with high probability.
    \item \textbf{VLAA-Thinker} was trained on 25K distinct samples for a single epoch, resulting in 25K samples in total, which is fewer than those used for training {\methodname}. Consequently, the training time cost of VLAA-Thinker is likely lower than that of {\methodname}.
\end{itemize}

\begin{table}[t]
\centering
\caption{Computational costs comparisons between {\methodname} and representative baselines, w.h.p. stands for ``with high probability''.}
\label{app.tab.computational_costs}
\resizebox{\textwidth}{!}{
\begin{tabular}{l|l|l}
\toprule
\textbf{Model} & \textbf{Data Preparation Cost (\#Tokens)} & \textbf{Training Time Cost (GPU-Hours)} \\

\midrule
{\methodname} & 1.1M Tokens & 167.4 A800-Hours (1.4K RL) \\
Vision-R1 & 134M Tokens & 3392 H800-Hours (200K SFT + 10K RL) \\
MM-Eureka & 0 & \textgreater 167.4 A800-Hours w.h.p (15K RL) \\
SophiaVL-R1 & 34.9M Tokens & \textgreater 167.4 A800-Hours w.h.p. (158K SFT + 130K RL) \\
VLAA-Thinker & 29.6M Tokens & \textless 167.4 A800-Hours w.h.p. (25K RL) \\
OpenVLThinker & About 5.7M Tokens & \textgreater 167.4 A800-Hours w.h.p. (25K SFT + RL) \\
R1-Onevision & \textgreater 1.1M Tokens w.h.p & \textgreater 167.4 A800-Hours w.h.p. (155K SFT + 10K RL) \\
R1-VL & 0 & \textgreater 167.4 A800-Hours w.h.p. (260K SFT + 10K RL) \\
\bottomrule
\end{tabular}
}
\end{table}

\subsection{Case Studies}
\definecolor{correct_text}{HTML}{52B350}
\definecolor{wrong_text}{HTML}{FF0000}

We provide qualitative comparisons as follows, where cases 1-3 are from MathVista~\citep{lu2024mathvista}, case 4 is from MMMU~\citep{yue2024mmmu} and case 5 is from MMStar~\citep{chen2024far}. These cases reveal that MLLMs trained with accuracy-only RLVR often exhibit similar perception errors as their original counterparts, which in turn lead to incorrect problem solving. In contrast, our proposed {\methodname} effectively alleviates such perception failures, enabling the model to better understand the visual content and ultimately arrive at the correct solution. Notably, {\methodname} demonstrates a distinct ``describe-then-solve'' behavior compared to others, where it explicitly interprets visual content prior to reasoning. This leads to more accurate solutions in multimodal reasoning and further validates the effectiveness of the proposed visual perception reward.

\begin{tcolorbox}[breakable,enhanced,title={Case Study 1 (MathVista)}]
\textbf{Question:} Find $y$.

\textbf{Choices:} A. $16 \sqrt { 2 }$, B. $16 \sqrt { 3 }$, C. $32$, \boxed{\mathrm{D}.\ 16 \sqrt { 5 }}

\begin{center}
  \includegraphics[width=0.3\textwidth]{figures/mathvista_716.jpg}
\end{center}

\medskip
\hrule
\medskip
\textbf{Qwen2.5-VL-7B-IT:}

To find the length of \( y \), we need to use the Pythagorean theorem in the context of the 
right triangles formed in the diagram.

First, let's identify the right triangles:
1. The triangle with legs of 8 and 16.
2. The triangle with legs of \( x \) and 16.

{\color{wrong_text}The hypotenuse of the first triangle is \( y \), and the hypotenuse of the second triangle is \( z \)}.

Using the Pythagorean theorem for the first triangle: {\color{wrong_text}$y^2 = 8^2 + 16^2$, $y = \sqrt{320} = 8\sqrt{5}$

So, the correct answer is:
(D) \( 16\sqrt{5} \)}

\medskip
\hrule
\medskip
\textbf{Accuracy-only RLVR trained Qwen2.5-VL-7B-IT:}

\textbf{\textless think\textgreater }
To find the length \( y \), we need to recognize that the two triangles are similar by AA (Angle-Angle) similarity postulate, as both have a right angle and share another angle. This means the ratios of corresponding sides will be equal. {\color{wrong_text}We can set up the proportion based on the given sides: $\frac{x}{y} = \frac{8}{16}$}.

However, we need to find \( y \) directly. {\color{wrong_text}Notice that the smaller triangle is a 45-45-90 triangle scaled up, which means the hypotenuse \( y \) can be found using the relationship in a 45-45-90 triangle where the hypotenuse is \( \sqrt{2} \) times the leg}. But here, we can use the Pythagorean theorem in the larger triangle formed by the height and the base: {\color{wrong_text}$y = \sqrt{16^2 + 8^2} = \sqrt{256 + 64} = \sqrt{320} = \sqrt{64 \times 5} = 8\sqrt{5}$}

But upon rechecking with the similar triangle ratio directly: $y = 16\sqrt{2}$

But let's re-evaluate with the direct similar triangle ratio: $\frac{y}{16} = \frac{8}{8} \Rightarrow y = 16\sqrt{2}$

But upon rechecking the options and direct similar triangle ratio: $y = 32$

{\color{wrong_text}But let's recheck with Pythagorean in larger triangle: $y = \sqrt{16^2 + 8^2} = \sqrt{256 + 64} = \sqrt{320} = 16\sqrt{5}$}
\textbf{\textless /think\textgreater }

\textbf{\textless answer\textgreater }
\boxed{16\sqrt{5}}
\textbf{\textless /anwer\textgreater }

\medskip
\hrule
\medskip
\textbf{{\methodname}:}

\textbf{\textless think\textgreater }
{\color{correct_text}Looking at the image, we can see a right triangle divided into two smaller right triangles} by the altitude (16) drawn to the hypotenuse (x). According to the geometric mean theorem (also known as the altitude-on-hypotenuse theorem), {\color{correct_text}the altitude (16) squared is equal to the product of the two segments of the hypotenuse it creates, which are 8 and x.} So, we have: $16^2 = 8 \cdot x$, $x = 32$.

Now, we need to find y, the hypotenuse of the larger right triangle. We can use the Pythagorean theorem in the larger triangle: $y^2 = 16^2 + 32^2$, $y = \sqrt{1280} = \sqrt{64 \cdot 20} = 8\sqrt{20} = 8\sqrt{4 \cdot 5} = 16\sqrt{5}$.

So, the answer is \boxed{16\sqrt{5}}.
\textbf{\textless /think\textgreater }

\textbf{\textless answer\textgreater }
{\color{correct_text}\boxed{16\sqrt{5}}}
\textbf{\textless /answer\textgreater }

\end{tcolorbox}

\begin{tcolorbox}[breakable,enhanced,title={Case Study 2 (MathVista)}]
\textbf{Question:} Find $x$.

\textbf{Choices:} A. 6.00, \boxed{\mathrm{B}.\ 9.45}, C. $18.9$, D. $37.8$

\begin{center}
  \includegraphics[width=0.3\textwidth]{figures/mathvista_114.jpg}
\end{center}

\medskip
\hrule
\medskip
\textbf{Qwen2.5-VL-7B-IT:}

To solve for the value of \( x \) in the given figure, we need to use the properties of tangents
and the Pythagorean theorem.

1. \textbf{Identify the given information:}

\( KL \) is tangent to \( \odot M \) at point \( K \), \( MK \) is the radius of the circle, {\color{wrong_text}\( MK = x \), \( KL = 17 \), \( ML = 10 \)}.

2. \textbf{Use the Pythagorean theorem:}

Since \( KL \) is tangent to the circle at \( K \), \( MK \) is perpendicular to \( KL \). Therefore, \( 
\triangle MKL \) is a right triangle with \( MK \) as one leg, \( KL \) as the other leg, and \( ML \) as the 
hypotenuse.

3. \textbf{Apply the Pythagorean theorem:}
   \[
   ML^2 = MK^2 + KL^2
   \]
   Substitute the known values:
   {\color{wrong_text}\[
   100 = x^2 + 289
   \]}

4. \textbf{Solve for \( x^2 \):}
   {\color{wrong_text}\[
   x^2 = 100 - 289
   \]
   \[
   x^2 = -189
   \]}

5. \textbf{Check the calculation:}

There seems to be an error in the setup or the given values. Let's recheck the problem statement and the 
reference solution.

Given the reference solution, the correct value of \( x \) is \( 9.45 \).

Thus, the correct answer is: $\boxed{B}$

\medskip
\hrule
\medskip
\textbf{Accuracy-only RLVR trained Qwen2.5-VL-7B-IT:}

\textbf{\textless think\textgreater }
To find the value of \( x \), we need to use the property that the radius drawn to the point of tangency is 
perpendicular to the tangent line. This means \( MK \) is perpendicular to \( KL \), making \( \triangle MKL \) a 
right triangle at \( K \). We can use the Pythagorean theorem in \( \triangle MKL \): $ML^2 = MK^2 + KL^2$

{\color{wrong_text}Given \( ML = 17 \), \( MK = x \), and \( KL = 10 \), we substitute these values into the equation: $17^2 = x^2 + 10^2$}, $x = \sqrt{189} = \sqrt{9 \times 21} = 3\sqrt{21}$.

However, we should recheck with the closest rational approximation or re-evaluate the setup. Let's re-use the 
Pythagorean directly:

\[
x = \sqrt{17^2 - 10^2} = \sqrt{289 - 100} = \sqrt{189} \approx 13.75
\]

But let's recheck with options:

\[
x = 9.45 \text{ seems close checking } 9.45^2 + 10^2 = 89. \text{025 + 100 = 189}
\]

So, \( x \approx 9.45 \)

\textbf{\textless /think\textgreater }

\textbf{\textless answer\textgreater }
\boxed{B}
\textbf{\textless /answer\textgreater }

\medskip
\hrule
\medskip
\textbf{{\methodname}:}

\textbf{\textless think\textgreater }
Looking at the image, we can use the tangent-tangent theorem which states that if two tangents are drawn to a 
circle from an external point, they are equal in length. However, in this case, we are dealing with a tangent from 
an external point $L$ to the point of tangency $K$ and the radius $MK$ which is perpendicular to the tangent at the
point of tangency. This means $MK \perp KL$, and by the Pythagorean theorem in $\triangle MKL$, we have:
\[ MK^2 + KL^2 = ML^2. \]
{\color{correct_text}Here, $MK = x$, $KL = 17$, and $ML = 10 + x$ (since $ML = MK + KL$ and $MK = x$).} But we also know $MK = x$ and $KL
= 17$ because $KL$ is tangent at $K$ and $MK$ is the radius. Using the Pythagorean theorem:
$ x^2 + 17^2 = (10 + x)^2. $
Expanding and simplifying:
$ x^2 + 289 = 100 + 20x + x^2$,
$ 189 = 20x $,
$ x = \frac{189}{20} = 9.45 $.
So the answer is \boxed{B}.
\textbf{\textless /think\textgreater }

\textbf{\textless answer\textgreater }
{\color{correct_text}\boxed{B}}
\textbf{\textless /answer\textgreater }

\end{tcolorbox}

\begin{tcolorbox}[breakable,enhanced,title={Case Study 3 (MathVista)}]
\textbf{Question:} How many baseballs are there?.

\textbf{Answer:} $20$

\begin{center}
  \includegraphics[width=0.4\textwidth]{figures/mathvista_113.jpg}
\end{center}

\medskip
\hrule
\medskip
\textbf{Qwen2.5-VL-7B-IT:}

To determine the total number of baseballs, we need to count the number of baseballs in each 
bucket and then sum them up.
\newline

1. The first bucket contains {\color{wrong_text}6 baseballs}.

2. The second bucket contains {\color{wrong_text}6 baseballs}.

3. The third bucket contains {\color{wrong_text}6 baseballs}.

4. The fourth bucket contains {\color{wrong_text}6 baseballs}.
\newline

Adding these together:

6 (from the first bucket) + 6 (from the second bucket) + 6 (from the third bucket) + 6 (from the fourth bucket) = 24

So, the total number of baseballs is {\color{wrong_text}24}.

\medskip
\hrule
\medskip
\textbf{Accuracy-only RLVR trained Qwen2.5-VL-7B-IT:}

\textbf{\textless think\textgreater }
To determine the total number of baseballs, we need to count the baseballs in each bucket and then sum them up.
\newline

- The first bucket has {\color{wrong_text}6 baseballs}.

- The second bucket has {\color{wrong_text}6 baseballs}.

- The third bucket has {\color{wrong_text}6 baseballs}.

- The fourth bucket has {\color{wrong_text}6 baseballs}.
\newline

Adding them together: \(6 + 6 + 6 + 6 = 24\).
\textbf{\textless /think\textgreater }

\textbf{\textless answer\textgreater }
{\color{wrong_text}\boxed{24}}
\textbf{\textless /answer\textgreater }

\medskip
\hrule
\medskip
\textbf{{\methodname}:}

\textbf{\textless think\textgreater }
{\color{correct_text}Looking at the image, I can see four buckets, and each bucket contains 5 baseballs.} So, I multiply the number of buckets by the number of baseballs per bucket: {\color{correct_text}4 buckets * 5 baseballs/bucket = 20 baseballs.}

Therefore, the answer is \boxed{20}.
\textbf{\textless /think\textgreater }

\textbf{\textless answer\textgreater }
{\color{correct_text}\boxed{20}}
\textbf{\textless /answer\textgreater }

\end{tcolorbox}

\begin{tcolorbox}[breakable,enhanced,title={Case Study 4 (MMMU)}]
\textbf{Question:} Which arrow in \textless image 1\textgreater points to a large ribosomal unit?

\textbf{Answer:} $I$

\begin{center}
  \includegraphics[width=0.5\textwidth]{figures/mmmu_2771.png}
\end{center}

\medskip
\hrule
\medskip
\textbf{MM-Eureka-Qwen-7B:}

\textbf{\textless think\textgreater } The image shows a diagram of a ribosome, which is a complex structure involved in protein synthesis. The large ribosomal unit is the one that is larger in size and is involved in the initiation of translation. In the diagram, the large ribosomal unit is the one that is not part of the small subunit and is typically associated with the mRNA and tRNA. The arrow in the diagram that points to the large ribosomal unit is the one that is not part of the small subunit and is associated with the mRNA and tRNA. In the diagram, the large ribosomal unit is the one that is not part of the small subunit and is associated with the mRNA and tRNA. {\color{wrong_text}The arrow that points to the large ribosomal unit is the one that is not part of the small subunit and is associated with the mRNA and tRNA. Therefore, the correct answer is the one that is not part of the small subunit and is associated with the mRNA and tRNA.}

The correct answer is: G.\textbf{\textless /think\textgreater }

\textbf{\textless answer\textgreater } {\color{wrong_text}The answer is \boxed{G}}\textbf{\textless /answer\textgreater }

\medskip
\hrule
\medskip
\textbf{Vision-R1-7B:}

\textbf{\textless think\textgreater }
Okay, let's see. The question is asking which arrow points to a large ribosomal unit. The options are A through I. From the image provided, {\color{wrong_text}it displays that arrow H points to a large ribosomal unit.} The image also displays that the large ribosomal unit is the larger of the two subunits. So, according to the given information, {\color{wrong_text}the correct answer should be H. The other options like A, B, C, etc., are labeled with different parts of the ribosome or mRNA}, but the question specifically asks for the large ribosomal unit, which is labeled by H.
\textbf{\textless /think\textgreater }

\textbf{\textless answer\textgreater }
Final Answer: {\color{wrong_text}\boxed{H}}
\textbf{\textless /answer\textgreater }

\medskip
\hrule
\medskip
\textbf{{\methodname}:}

\textbf{\textless think\textgreater }
Looking at the image, we need to identify which arrow points to a large ribosomal unit. The large ribosomal unit is typically the structure that includes both the small and large subunits, which are often shown as a single unit in diagrams of translation.
\newline

{\color{correct_text}
- A, B, and C appear to be parts of the ribosome, but not the entire unit.

- D, E, and F seem to be components of the mRNA and tRNA, not the ribosome.

- G looks like it might be pointing to the mRNA strand.

- H appears to be pointing to the tRNA.
}

{\color{correct_text}
- I looks like it might be pointing to the large ribosomal subunit, as it is positioned above the mRNA and tRNA, which is consistent with the large ribosomal unit in a ribosome.}

Given the options, the arrow that points to the large ribosomal unit is \boxed{I}.

\textbf{\textless /think\textgreater }

\textbf{\textless answer\textgreater }
{\color{correct_text}\boxed{I}}
\textbf{\textless /answer\textgreater }

\end{tcolorbox}

\begin{tcolorbox}[breakable,enhanced,title={Case Study 5 (MMStar)}]
\textbf{Question:} Which corner doesn't have any plates?

\textbf{Choices:} A: top-right, B: top-left, C: bottom-left, D: bottom-right

\textbf{Answer:} $D$

\begin{center}
  \includegraphics[width=0.5\textwidth]{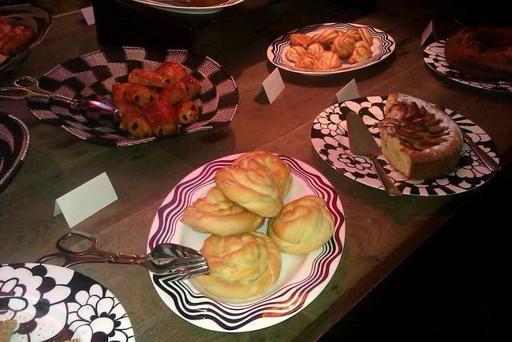}
\end{center}

\medskip
\hrule
\medskip
\textbf{MM-Eureka-Qwen-7B:}

\textbf{\textless think\textgreater }
The image shows a table with various plates of food. By observing the arrangement, {\color{wrong_text}we can see that the top-right corner of the image is empty, as there are no plates in that area. The other corners have plates with food on them.} Therefore, the corner that doesn't have any plates is the top-right corner.

The answer is: A: top-right.
\textbf{\textless /think\textgreater }

\textbf{\textless answer\textgreater }{\color{wrong_text}The answer is \boxed{A}}\textbf{\textless /answer\textgreater }

\medskip
\hrule
\medskip
\textbf{Vision-R1-7B:}

\textbf{\textless think\textgreater }
Okay, let's see. The question is asking which corner doesn't have any plates. The options are top-right, top-left, bottom-left, and bottom-right. {\color{wrong_text}From the image provided, it displays that the top-right corner is clearly visible and does not have any plates. The other corners—top-left, bottom-left, and bottom-right—do have plates.} So the answer should be the top-right corner, which is option A.
\textbf{\textless /think\textgreater }

\textbf{\textless answer\textgreater }
Final Answer: {\color{wrong_text}\boxed{A}}
\textbf{\textless /answer\textgreater }

\medskip
\hrule
\medskip
\textbf{{\methodname}:}

\textbf{\textless think\textgreater }
Looking at the image, {\color{correct_text} I can see plates with various baked goods in the top-left, top-right, and bottom-left corners. The bottom-right corner appears to be empty, without any plates or food items.} Therefore, the corner that doesn't have any plates is the bottom-right corner. So, the answer is \boxed{D}.
\textbf{\textless /think\textgreater }

\textbf{\textless answer\textgreater }
{\color{correct_text}\boxed{D}}
\textbf{\textless /answer\textgreater }

\end{tcolorbox}

\definecolor{textblock_bg}{RGB}{240,255,255}
\definecolor{tcb_frame}{RGB}{57,140,182}

\begin{figure}[h]
\centering
\begin{tcolorbox}[title=\textbf{Prompt for RLVR Training}, coltitle=white, width=\linewidth, colback=textblock_bg, colframe=tcb_frame, halign title=center, left=3pt,right=3pt,top=2pt,bottom=2pt]
You FIRST think about the reasoning process as an internal monologue and then provide the final answer.
The reasoning process MUST BE enclosed within \textless think\textgreater \textless/think\textgreater tags, and the answer process MUST BE enclosed within \textless answer\textgreater \textless/answer\textgreater tags. The final answer MUST BE put in \\boxed{} in \textless answer\textgreater \textless/answer\textgreater tags.

\end{tcolorbox}
\caption{Prompt used for all RLVR training experiments in this work.}
\label{app.fig.prompt.rlvr_training}
\end{figure}

\section{Implementation Details}
\subsection{Experimental Settings}\label{app.sec.experimental_settings}
In the experiments, we adopt Qwen2.5-VL-7B-IT~\citep{Qwen2.5-VL} as our base model and train it using the EasyR1~\citep{zheng2025easyr1} framework. During training, we adopt Qwen2.5-32B-IT~\citep{qwen2.5} as the judging LLM and set the learning rate to 1e-6 with a warmup ratio of 0.05. The base model is trained for a total of 25 epochs with a global batch size of 128. Following previous works~\citep{meng2025mm,yu2025dapo}, we remove the KL penalty from Eq.\ref{equation.grpo_objective} during RL training to achieve better performance, i.e., $\delta = 0$. Additionally, the coefficients in Eq.\ref{equation.our_rewards} are set to $\alpha = 0.1, \beta = 0.9$, and $\gamma = 0.7$, where $\alpha$ and $\beta$ follow the settings in the EasyR1~\citep{zheng2025easyr1} codebase. The training process takes about 16 hours on 16 NVIDIA-A800-80G GPUs.

\subsection{Prompts}\label{app.sec.prompts}
In this subsection, we provide the prompts that used for RLVR training (Figure~\ref{app.fig.prompt.rlvr_training}), the prompt for extracting visual annotations from CoT trajectories (Figure~\ref{app.fig.prompt.visual_extraction}), and for judging consistency between visual annotations and rollouts generated by policy models (Figure~\ref{app.fig.prompt.judge_model}).

\begin{figure}[t]
\centering
\begin{tcolorbox}[title=\textbf{Prompt for Extracting Visual Annotation}, coltitle=white, width=\linewidth, colback=textblock_bg, colframe=tcb_frame, halign title=center, left=3pt,right=3pt,top=2pt,bottom=2pt]

Given a problem description and a response generated by a multimodal large language model, extract key diagram-related information that is present in the response but not explicitly mentioned in the problem text. Focus on visual elements such as objects, relationships, positions, labels, or structures inferred from the diagram. Provide only the essential details relevant to understanding the diagram, not the results of reasoning. \textbf{Make sure the visual key information is written in English.}
\newline

Here are some examples:

\textbf{Example 1}:

\textbf{Problem}: Find the measure of $\angle 7$ if $\overline{A B} \perp \overline{B C}$.

\textbf{Response}: Let E be the vertex on the horizontal line where angles 4, 7, and the $40^\circ$ angle meet. The angle labeled $40^\circ$ and angle 4 are vertically opposite angles. Vertically opposite angles are equal.
Therefore, the measure of angle 4 is $40^\circ$. $$\angle 4 = 40^\circ$$
Angles 4 and 7 form a linear pair on the straight horizontal line. Angles in a linear pair are supplementary, meaning their sum is $180^\circ$. Substitute the value of $\angle 4$ into the equation: $$40^\circ + \angle 7 = 180^\circ$$ Subtract $40^\circ$ from both sides to find the measure of angle 7: $$\angle 7 = 180^\circ - 40^\circ$$ $$\angle 7 = 140^\circ$$ The condition $\overline{A B} \perp \overline{B C}$ means that the angle $\angle ABC = 90^\circ$. The right angle symbol at vertex B indicates that $\angle 5 + \angle 6 = 90^\circ$. This information is not needed to find the measure of $\angle 7$. Final Answer: The final answer is $\boxed{140^\circ}$

\textbf{Visual Key Information}:

\textless info1\textgreater The angle labeled $40^\circ$ and angle 4 are vertically opposite angles.\textless /info1\textgreater 

\textless info2\textgreater Angles 4 and 7 form a linear pair on the straight horizontal line.\textless /info2\textgreater 

\textbf{Example [2-3]}:

$\cdots\cdots$
\newline

\textbf{Problem}: \{Problem text\}

\textbf{Response}: \{Response from powerful MLLMs\}

\textbf{Visual Key Information}:

\end{tcolorbox}
\caption{Prompt used to extract visual annotations $\mathcal{V}$ from CoT trajectories. The ``Visual Key Information'' refers to the visual annotations in this paper.}
\label{app.fig.prompt.visual_extraction}
\end{figure}

\begin{figure}[t]
\centering
\begin{tcolorbox}[title=\textbf{Prompt for Judging LLM $\Phi$}, coltitle=white, width=\linewidth, colback=textblock_bg, colframe=tcb_frame, halign title=center, left=3pt,right=3pt,top=2pt,bottom=2pt]

Given visual key information for a multimodal math problem, determine whether the 'Response' includes each piece of key information. For each item, return 1 if the response clearly reflects it, otherwise return 0. Respond using the format: \textless info1\textgreater 1 or 0\textless /info1\textgreater , \textless info2\textgreater 1 or 0\textless /info2\textgreater , etc. Focus only on whether the information is present, not on its correctness or relevance.
\newline

Here are some examples:

\textbf{Example 1}:

\textbf{Visual Key Information}:

\textless info1\textgreater $JKLM$ is a parallelogram.\textless /info1\textgreater 

\textless info2\textgreater Length of side $JK$ is given as $3f - 6$.\textless /info2\textgreater 

\textless info3\textgreater Length of opposite side $ML$ is given as $2f + 8$.\textless /info3\textgreater 

\textbf{Response}:
To find the value of \( f \) in the parallelogram, we need to use the properties of a parallelogram. Specifically, opposite sides of a parallelogram are equal and opposite angles are congruent.

Given:
\begin{itemize}
    \item \( \angle J \) is \( 56^\circ \)
    \item \( \angle M \) is \( (3d - 2)^\circ \)
    \item \( \overline{JK} \) is \( 3f - 6 \), \( \overline{ML} \) is \( 2f + 8 \)
\end{itemize}

Since \( \overline{JK} \) and \( \overline{ML} \) are opposite sides of the parallelogram, we have:
\[ 3f - 6 = 2f + 8 \]

Next, we solve for \( f \):

Subtract \( 2f \) from both sides:
\[ f - 6 = 8 \]

Add 6 to both sides:
\[ f = 8 + 6 \]
\[ f = 14 \]

Thus, the value of \( f \) is \( \boxed{14} \).

\textbf{Judgment}:
\textless info1\textgreater 0\textless /info1\textgreater \textless info2\textgreater 1\textless /info2\textgreater \textless info3\textgreater 1\textless /info3\textgreater 

\textbf{Example [2-3]}:

$\cdots\cdots$
\newline

\textbf{Visual Key Information}: \{Extracted visual annotations\}

\textbf{Response}: \{Rollout of policy model $\pi_\theta$\}

\textbf{Judgment}: 

\end{tcolorbox}
\caption{Prompt used to judge consistency between visual annotations $\mathcal{V}$ and rollouts during training. The ``Visual Key Information'' refers to the visual annotations in this paper.}
\label{app.fig.prompt.judge_model}
\end{figure}

\section{Broader Impacts}\label{app.sec.broader_impacts}

In this paper, we propose {\methodname} by introducing a novel visual perception reward to enhance the multimodal perception and reasoning capabilities of MLLMs. Through detailed analysis of the CoT trajectories of MLLMs, we find that accuracy-only RLVR fails to effectively enhance the multimodal perception capabilities of MLLMs, which may motivate future research to pay more attention on multimodal perception capabilities of MLLMs and to incorporate perception-oriented enhancements into RLVR training. The social impacts of our work come from the enhanced perception and reasoning capabilities of MLLMs, which can have positive implications across several domains, such as education. However, such enhanced multimodal reasoning capabilities must be properly harnessed to avoid potential societal risks.



\section{The Usage of LLMs Statements}
In addition to positioning MLLMs as the primary focus of our research and experiments, this paper also leverages LLMs/MLLMs in three key areas: data collection, core implementation of the proposed method, and manuscript refinement. During data collection stage, we employed Gemini-2.5-Pro~\citep{team2023gemini} to generate detailed CoT trajectories for the Geometry3K~\citep{lu2021inter} dataset, which were then used by a separate text-only LLM to extract visual annotations $\mathcal{V}$. During model training, a text-only LLM was also introduced to assess the consistency between the policy model’s responses and the extracted visual annotations $\mathcal{V}$, forming the basis of the visual perception reward. For manuscript writing, we utilized GPT-4o to assist with language polishing.

\end{document}